\documentclass[acmsmall,screen]{acmart}
\AtBeginDocument{%
  }

\setcopyright{acmlicensed}
\copyrightyear{2026}
\acmYear{2026}
\acmDOI{XXXXXXX.XXXXXXX}

\acmJournal{CSUR}
\renewcommand\footnotetextcopyrightpermission[1]{}
\usepackage[utf8]{inputenc} % allow utf-8 input
\usepackage[T1]{fontenc}    % use 8-bit T1 fonts
\usepackage{url}            % simple URL typesetting
\usepackage{color, soul}
\usepackage{wrapfig}

\usepackage{tabularx}
\usepackage{ragged2e}
\usepackage{multirow}
\usepackage{booktabs}       % professional-quality tables
\usepackage{amsfonts}       % blackboard math symbols
\usepackage{nicefrac}       % compact symbols for 1/2, etc.
\usepackage{microtype}      % microtypography
\usepackage{graphicx}
\usepackage{csquotes}
\usepackage{amsmath}
\usepackage{enumitem}       
\usepackage{blindtext} %This package generates automatic text
\usepackage{epigraph} 
\definecolor{deepblue}{RGB}{0,51,160}
\definecolor{diffblue}{RGB}{0,90,200}
\newcommand{\add}[1]{#1}
\usepackage{booktabs}
\usepackage{tabularx}
\usepackage[table]{xcolor}
\usepackage{ragged2e}
\usepackage{makecell}     % for manual line breaks in a cell
\newcolumntype{Y}{>{\RaggedRight\arraybackslash}X}

\usepackage{fontawesome}
\usepackage{pifont}
\usepackage{adjustbox}
\usepackage{tikz}
\usepackage{pgfplots}
\pgfplotsset{compat=1.18}
\usepackage[edges]{forest}
\definecolor{framework-blue}{RGB}{47, 85, 151}

\definecolor{content-yellow}{RGB}{255, 230, 153}
\definecolor{framework-yellow}{RGB}{255, 255, 255}
\definecolor{content-orange}{RGB}{251, 229, 215}
\definecolor{framework-orange}{RGB}{248, 203, 175}
\definecolor{content-gray}{RGB}{237, 237, 237}
\definecolor{framework-gray}{RGB}{166, 166, 166}

\definecolor{paired-light-blue}{RGB}{198, 219, 239}
\definecolor{paired-dark-blue}{RGB}{49, 130, 188}
\definecolor{paired-light-orange}{RGB}{251, 208, 162}
\definecolor{paired-dark-orange}{RGB}{230, 85, 12}
\definecolor{paired-light-green}{RGB}{199, 233, 193}
\definecolor{paired-dark-green}{RGB}{49, 163, 83}
\definecolor{paired-light-purple}{RGB}{218, 218, 235}
\definecolor{paired-dark-purple}{RGB}{117, 107, 176}
\definecolor{paired-light-gray}{RGB}{217, 217, 217}
\definecolor{paired-dark-gray}{RGB}{99, 99, 99}
\definecolor{paired-light-pink}{RGB}{222, 158, 214}
\definecolor{paired-dark-pink}{RGB}{123, 65, 115}
\definecolor{paired-light-red}{RGB}{231, 150, 156}
\definecolor{paired-dark-red}{RGB}{131, 60, 56}
\definecolor{paired-light-yellow}{RGB}{231, 204, 149}
\definecolor{paired-dark-yellow}{RGB}{141, 109, 49}
\tikzset{%
    parent/.style = {align=center,text width=2.5cm,rounded corners=3pt, line width=0.3mm, fill=gray!10,draw=gray!80},
    child/.style = {align=center,text width=2.3cm,rounded corners=3pt, fill=blue!10,draw=blue!80,line width=0.3mm},
    grandchild/.style = {align=center,text width=2cm,rounded corners=3pt},
    greatgrandchild/.style = {align=center,text width=1.5cm,rounded corners=3pt},
    greatgrandchild2/.style = {align=center,text width=1.5cm,rounded corners=3pt},    
    referenceblock/.style =  {align=center,text width=1.5cm,rounded corners=2pt},
    brain/.style = {align=center,text width=2.2cm,rounded corners=3pt, fill=white,draw=framework-blue,line width=0.3mm},   
    brain_work/.style = {align=center, text width=4.5cm,rounded corners=3pt, fill=white,draw=framework-blue,line width=0.3mm},
    perception/.style= {align=center,text width=2.2cm,rounded corners=3pt, fill=white,draw=framework-blue,line width=0.3mm},
    perception_work/.style= {align=center, text width=4.5cm,rounded corners=3pt, fill=white,draw=framework-blue,line width=0.3mm}, 
    action/.style= {align=center,text width=2.2cm,rounded corners=3pt, fill=white,draw=framework-blue,line width=0.3mm},
    action_work/.style= {align=center, text width=4.5cm,rounded corners=3pt, fill=white,draw=framework-blue,line width=0.3mm},
    single_agent/.style= {align=center,text width=2.2cm,rounded corners=3pt, fill=white,draw=framework-blue,line width=0.3mm},
    single_agent_work/.style= {align=center, text width=4.5cm,rounded corners=3pt, fill=white,draw=framework-blue,line width=0.3mm},
    multi_agent/.style= {align=center,text width=2.2cm,rounded corners=3pt, fill=white,draw=framework-blue,line width=0.3mm},
    multi_agent_work/.style= {align=center, text width=4.5cm,rounded corners=3pt, fill=white,draw=framework-blue,line width=0.3mm},
    human_agent/.style= {align=center,text width=2.2cm,rounded corners=3pt, fill=white,draw=framework-blue,line width=0.3mm},
    human_agent_work/.style= {align=center, text width=4.5cm,rounded corners=3pt, fill=white,draw=framework-blue,line width=0.3mm},
    behavior_and_personality/.style= {align=center,text width=2.2cm,rounded corners=3pt, fill=white,draw=framework-blue,line width=0.3mm},
    behavior_and_personality_work/.style= {align=center, text width=4.5cm,rounded corners=3pt, fill=white,draw=framework-blue,line width=0.3mm},
    society_environment/.style= {align=center,text width=2.2cm,rounded corners=3pt, fill=white,draw=framework-blue,line width=0.3mm},
    society_environment_work/.style= {align=center, text width=4.5cm,rounded corners=3pt, fill=white,draw=framework-blue,line width=0.3mm},
    society_simulation/.style= {align=center,text width=2.2cm,rounded corners=3pt, fill=white,draw=framework-blue,line width=0.3mm},
    society_simulation_work/.style= {align=center, text width=4.5cm,rounded corners=3pt, fill=white,draw=framework-blue,line width=0.3mm},
}

\begin{document}

\title{A Survey on Large Language Model-Based Game Agents}

\author{Sihao Hu}
\email{sihaohu@gatech.edu}
% \orcid{0000-0003-3297-6991}
\affiliation{
  \institution{Georgia Institute of Technology}
   \country{USA}
}

\author{Tiansheng Huang}
\email{thuang@gatech.edu}
% \orcid{0000-0002-4557-1865}
\affiliation{
  \institution{Georgia Institute of Technology}
  \country{USA}
}

\author{Gaowen Liu}
% \orcid{0009-0007-9048-9274}
\email{gaoliu@cisco.com}
\affiliation{%
  \institution{Cisco Research}
    \country{USA}
}

\author{Ramana Rao Kompella}
% \orcid{0000-0002-7559-8997}
\email{rkompell@cisco.com}
\affiliation{%
  \institution{Cisco Research}
    \country{USA}
}

\author{Fatih Ilhan}
\email{filhan@gatech.edu}
% \orcid{0000-0002-0173-7544}
\affiliation{%
  \institution{Georgia Institute of Technology}
    \country{USA}
}

\author{Selim Furkan Tekin}
% \orcid{0000-0002-8662-3609}
\email{stekin6@gatech.edu}
\affiliation{%
  \institution{Georgia Institute of Technology}
    \country{USA}
}

\author{Yichang Xu}
% \orcid{0009-0000-1094-4206}
\email{xuyichang@gatech.edu}
\affiliation{%
  \institution{Georgia Institute of Technology}
    \country{USA}
}

\author{Zachary Yahn}
% \orcid{0000-0003-2408-2576}
\email{zachary.yahn@gatech.edu}
\affiliation{%
  \institution{Georgia Institute of Technology}
    \country{USA}
}

\author{Ling Liu}
% \orcid{0000-0002-4138-3082}
\email{ling.liu@cc.gatech.edu}
\affiliation{%
  \institution{Georgia Institute of Technology}
    \country{USA}
}

\renewcommand{\shortauthors}{Hu et al.}

\begin{abstract}

Game environments provide rich, controllable settings that simulate many aspects of real-world complexity. As such, game agents offer a valuable testbed for exploring capabilities relevant to Artificial General Intelligence~\cite{Yannakakis2018AIGames}. Recently, the emergence of Large Language Models (LLMs) provides new opportunities to endow these agents with generalizable reasoning, memory, and adaptability in complex game environments. 
This survey offers an up-to-date review of LLM-based game agents (LLMGAs) through a unified reference architecture. At the single-agent level, we synthesize existing studies around three core components: memory, reasoning, and perception–action interfaces, which jointly characterize how language enables agents to perceive, think, and act. 
At the multi-agent level, we outline how communication protocols and organizational models support coordination, role differentiation, and large-scale social behaviors. 
To contextualize these designs, we introduce a challenge-centered taxonomy linking six major game genres to their dominant agent requirements, from low-latency response in action games to open-ended goal formation in sandbox worlds. We will continuously update the survey to track new developments, and we maintain a continuously updated list of related papers at: \url{https://github.com/git-disl/awesome-LLM-game-agent-papers}.

\end{abstract}



% \keywords{Large language models, game agents, LLM-based agents, multi-agent systems, game AI}

% \keywords{Large language models, game agents, multi-agent systems, game AI}


\maketitle

% \newpage
%{
%  \hypersetup{linkcolor=RoyalBlue, linktoc=page}
  %\hypersetup{linkcolor=black, linktoc=section}
%  \tableofcontents
%}
% \clearpage

% \vspace{-0.1cm}
\section{Introduction}

By scaling model capacity and training on massive, diverse text corpora, large language models (LLMs) have demonstrated strong capabilities in language understanding, knowledge generalization, and conversational dialogue~\cite{InstructGPT,GPT3,GPT4}. Despite these advances, current LLMs are primarily optimized on fixed, static text corpora. Human intelligence, in contrast, develops through continuous sensorimotor engagement with the environment~\cite{smith2005development},  
for example, by forming perceptual representations from repeated interactions that capture the structure and dynamics of the world~\cite{barsalou1999perceptual},  
and by adjusting behavior in response to feedback from action outcomes that gradually improves performance~\cite{clark2013whatever}. In general, the literature on embodied cognition emphasizes that human intelligence arises from situated interaction with the environment rather than from disembodied symbol manipulation~\cite{clark1998being,varela2017embodied,smith2005development}.

Unlike humans, LLM-based agents lack a physical body, making deep participation in real-world interactions difficult and costly. In contrast, game environments provide a natural testbed for realizing the coupling between agent and environment, and offer a richer, more embodied alternative compared to typical settings of current LLM-based agents, such as dialogue, web navigation, or API tool use~\cite{wang2023survey}. By granting avatars to agents in the interactive world with perception and action modules, digital games approximate aspects of real-world while remaining safe, controllable, and cost-effective. In addition, they are reproducible and span a wide range of complexity, making them an effective platform for advancing LLMs toward interactive intelligence.

Traditional game agents follow a control-based paradigm, where decision-making is coupled through predefined or learned state–action mappings~\cite{Yannakakis2018AIGames}. Finite state machines, behavior trees, and reinforcement learning agents~\cite{iovino2022survey,sutton1998reinforcement} exemplify this design. In contrast, language serves as a unified medium for LLM-based agents to represent goals, contexts, and interactions, enabling explicit reasoning, reflection, and communication beyond traditional systems. 

% , leaving the field of LLM-based game agents (LLMGAs) underexplored.

Existing surveys~\cite{wang2023survey,gao2023large} touch on the topic from different angles yet largely treat games as one of downstream applications alongside dialogue, tool use, or web automation. However, the complexity and openness of game environments distinguish them from narrowly defined tasks. For instance, while a web-based agent may complete a query or transaction through a handful of API calls, a sandbox game enables researchers to cultivate entire agent societies and allows agents to freely explore, interact, and build within physics-driven worlds. These game environments afford a degree of freedom that enables emergent behaviors far beyond constrained, task-oriented interactions. On the other hand, game-focused surveys~\cite{gallotta2024large,sweetser2024large} emphasize areas such as game development, educational applications, or content generation, leaving the field of LLM-based game agents (LLMGAs) underexplored. As a result, a dedicated survey of LLMGAs as a distinct research area is in strong demand.

To bridge this gap, this survey focuses exclusively on LLM-based 
game agents. Our contributions include (i) a reference architecture for analyzing LLMGAs at both single-agent and multi-agent levels, and (ii) a challenge-centered game taxonomy linking game genres to agent design requirements. Through these two lenses, we review existing studies and identify open challenges and future directions. The remainder of this paper is organized as follows. Section~\ref{sec:research_method} describes our research methodology. Section~\ref{sec:overview} provides an overview of the LLMGA framework and game taxonomy. Section~\ref{sec:memory}, Section~\ref{sec:reasoning}, and Section~\ref{sec:in/output} detail the three core single-agent components: memory system, reasoning mechanism, and perception-action interface. Section~\ref{sec:multi_agent_system} covers the multi-agent framework. Section~\ref{sec:taxonomy} applies the challenge-centered taxonomy to analyze design challenges and methods. Section~\ref{sec:discussion} synthesizes the reviewed literature and identifies the open challenges. Section~\ref{sec:conclusion} concludes the survey.

\vspace{-0.2cm}
\section{Research Methodology}
\label{sec:research_method}

% This section presents the research methodology of this survey, including the research objectives and questions, and the source retrieval process.

% \vspace{-0.2cm}
\subsection{Research Objectives and Questions}

This survey is guided by two main objectives. The first objective is to clarify the architectural landscape of LLM-based agents in games and identify common design patterns across existing studies. The second objective is to analyze how contextual characteristics of games, such as game genre, relate to architectural design decisions in LLMGAs. Based on these objectives, this survey addresses the following research questions: 

\textit{(Q1) How are the core architectural components of LLMGA frameworks designed and implemented in existing studies?} 

\textit{(Q2) How do different game genres influence architectural design requirements in LLMGAs?}

To answer Q1, we categorize existing LLMGA studies under a unified reference architecture that integrates two complementary perspectives. The first, the LLMGA framework, enables component-level analysis of a single agent. It abstracts common design choices into three modules: a memory system that stores and retrieves past experience, a reasoning mechanism that plans and makes decisions, and a perception \& action interface that connects the agent to the game environment. The second, the multi-LLMGA framework, examines how populations of agents interact and self-organize. It distinguishes two levels: agent-level communication, which governs how agents exchange messages and align their beliefs, and organization-level structure, which shapes how agents are coordinated and organized collectively.

To answer Q2, we introduce a challenge-centered taxonomy that maps six representative game genres~\cite{steamdb2025,lee2014facet} to the distinct demands they impose on agent design. For example, role-playing games center on the problem of role fidelity, \textit{i.e.}, how to encode and maintain consistent personas in memory so that dialogue and actions remain aligned with character identity over extended interactions. These genre–challenge mappings offer a structured lens on prior work and practical guidance for developing future LLMGAs. The broader aim of this survey is to position game environments as experimental grounds for examining whether sustained interaction between agents and their environments can foster more general and adaptive forms of intelligence.

% \vspace{-1.0em}
% \centering
% \includegraphics[width=4.8cm]{./Figure/action_game.png}
% \vspace{-0.7cm}
% \caption{Action games require agents to respond with low latency and execute precise low-level control.}
% \label{fig:action_game}
% \vspace{-1.0em}

\subsection{Source Retrieval Process}

% \begin{wrapfigure}{r}{5cm}

We include papers that employ an LLM or vision language model (VLM) as a central decision-making component and  involve interaction with a game or game-like environment. The source retrieval process can be divided into three steps:
In the first step, we conducted keyword searches with the Boolean query (“large language model” OR LLM) AND (game OR gaming) AND (agent) across four sources, including ACM Digital Library, IEEE Xplore, Google Scholar, and arXiv, covering publications from January 1, 2022 to June 1, 2026.  In the second step, we used Rayyan\footnote{Rayyan is a web-based tool for collaborative screening in systematic reviews, supporting blinded inclusion/exclusion decisions. \url{https://www.rayyan.ai/}} to facilitate collaborative screening by six co-authors. Each co-author independently reviewed titles and abstracts in blind mode and labeled each record as included or excluded. Papers receiving at least two independent include votes were retained. We then conducted full-text screening by evenly assigning the papers to the same six co-authors, who labeled each paper as included or excluded based on its relevance to the survey scope. In the third step, we conducted backward snowballing by reviewing the recent references on LLMGAs. Identified candidates were subjected to the same collaborative full-text screening described in the second step, resulting in a final corpus analyzed in this survey.

\section{Overview}
\label{sec:overview}

% Traditional game agents follow a control-based paradigm, where the agent observes the environment, selects a behavior, and executes it within a predefined task space~\cite{Yannakakis2018AIGames}. Methods such as finite state machines (FSMs), behavior trees (BTs), and traditional Reinforcement Learning agents~\cite{iovino2022survey,shao2019survey,sutton1998reinforcement} embody this principle in various forms, from rule-based control to learned policies. Despite their differences, perception and decision remain tightly coupled through state–action mappings, with reasoning and memory, if present, embedded implicitly within the control logic or learned policy parameters. In contrast, LLM-based game agents operate through language, which serves as a general medium for representing goals, contexts, and interactions, and enables explicit reasoning, reflection, and communication, allowing agents to explain, justify, and coordinate their behavior in ways traditional systems cannot. 

\subsection{LLM-based Game Agent (LLMGA) Framework}

\begin{figure}
    \centering
    % \vspace{-0.4cm}
    \includegraphics[width=14cm]{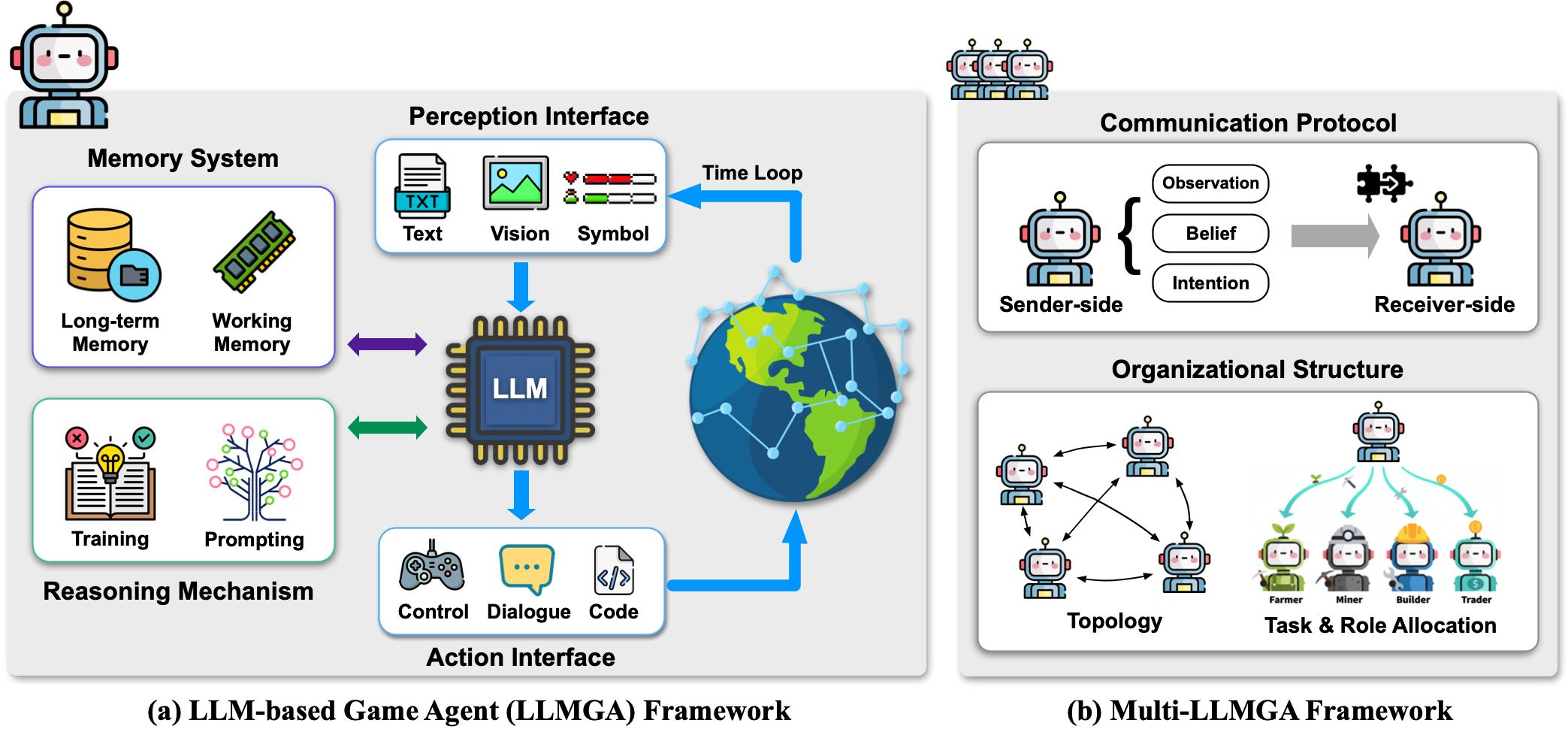}
    \vspace{-0.5cm}
    \caption{
    (a) Single-agent framework for LLMGAs, consisting of a memory system, a reasoning mechanism, and interfaces for perception and action. These modules are connected through the central LLM, driving a continuous gameplay loop where the agent perceives the evolving environment and acts in response. 
    (b) Multi-LLMGA framework that extends the architecture to populations of agents, including the communication protocol that governs message exchange and the organizational structure that determines topology, task allocation, and role differentiation.
    }
    \label{fig:game_agent_framework}
    \vspace{-0.4cm}
\end{figure}

Cognitive science views intelligence as an integrated system in which perception–action, memory, and reasoning processes interact to produce adaptive behavior~\cite{newell1994unified,kotseruba202040}. In line with this view, we find that existing studies on LLMGAs primarily introduce techniques that fall into three components: memory, reasoning, and perception–action~\cite{park2023generative,yao2022react,hu2024pokellmon}. Building on this perspective, we categorize existing LLMGA studies under a unified framework that instantiates these cognitive principles through the three components. Figure~\ref{fig:game_agent_framework}(a) illustrates the overall architecture: a central LLM connects the three components in continuous interaction with the game environment. At each step of gameplay, the environment evolves and produces new observations, which the agent perceives, interprets, and acts upon, completing a closed perception-action loop.

The \textbf{perception interface} transforms these observations into representations that the LLM can interpret~\cite{starcraftii}. In Section~\ref{sec:in/output}, we discuss how different modalities of observations, including textual, symbolic, and visual inputs, are handled by the agent.

The \textbf{memory system} provides a temporal mechanism that links past, present, and future, allowing information to persist across time and guide ongoing decisions. Following classic distinctions in cognitive psychology~\cite{BaddeleyHitch1974,baddeley2012working}, we divide it into working memory and long-term memory. Working memory offers a short-term buffer that supports immediate processing and coordination across steps, with technical considerations centered on extending its capacity and maintaining consistency over time. Long-term memory, by contrast, accumulates knowledge and experience across episodes. In Section~\ref{sec:memory}, we will focus on how to decide when and what to consolidate from transient experiences into long-term memory, and how stored content can be structured and retrieved.

Building on observations and memories, the \textbf{reasoning mechanism} defines how the LLM generates reasoning traces, such as plans, explanations, or self-critiques, that guide action proposals~\cite{CoT,yao2022react,shinn2023reflexion}. In cognitive science, reasoning is understood as constructing and operating on internal representations to draw inferences beyond the given information~\cite{johnson2010mental,evans2008dual}. In Section~\ref{sec:reasoning}, we outline two complementary approaches: prompting strategies, which elicit diverse reasoning paths at inference time, ranging from single linear chains to multiple parallel explorations and iterative refinements; Training paradigms, which improve reasoning ability by learning from expert demonstrations and from trial-and-error interaction with the environment.

Finally, the \textbf{action interface} functions as the agent’s hand and foot, translating language-based action proposals into concrete interactions with the environment~\cite{wang2023voyager}. In Section~\ref{sec:in/output}, we discuss how high-level, free-form language decisions are transformed into executable behaviors, including constrained natural language commands, symbolic actions, and sequences of low-level controls. These actions in turn alter the game state, producing new observations and completing the cycle of interaction.

\subsection{Multi-LLMGA Framework}

% Building on the single-agent framework, the multi-agent framework introduces an additional layer of complexity: agents not only interact with the environment but also with each other. 
% Such settings naturally call for mechanisms of coordination and communication~\cite{hutchins1995cognition,wooldridge2009introduction}.  Compared to generic LLM-based multi-agent systems, game environments impose additional constraints such as partial observability, limited communication bandwidth, and the need to preserve realistic gameplay boundaries, making their design challenges distinct. 

In games, agents interact not only with the environment but also with one another, which naturally calls for explicit mechanisms for coordination and communication. Game environments impose realistic constraints on information sharing: observations are distributed across agents, communication channels are often bandwidth-limited, and direct sharing of internal states or memories is typically disallowed or time-constrained. In such environments, communication between agents needs to be explicitly designed to operate under constrained channels~\cite{zhang2023building}. \textbf{Communication protocols} therefore formalize what information is shared, when it is shared, and how it is interpreted by the receiver~\cite{qian2025scaling}. Without such protocols, directly transmitting raw observations is often inefficient, noisy, and prone to conflict with an agent’s local beliefs.

When multiple agents act in a shared environment, decision conflicts, redundant actions, and inconsistent plans become unavoidable. We need explicit mechanisms to manage these challenges, such as introducing decision authority and responsibilities. \textbf{Organizational structures} are therefore necessary to manage, track and constrain how decisions are made and combined~\cite{li2024survey,qian2025scaling}. By defining role assignments and coordination pathways, organizational structures reduce coordination complexity and prevent uncontrolled interaction among agents. Such organizational structures help prevent situations in which coordination overhead grows combinatorially with the number of agents, thereby reducing inefficiency in cooperative behavior.

% At the agent level, the \textbf{communication protocol} specifies how information flows between peers and how it is integrated into ongoing cognition. Directly transmitting raw observations is often overwhelming and noisy. Therefore, messages should be filtered and abstracted into higher-level forms such as beliefs or intentions. Upon receiving a message, an agent must reconcile the new content with its own memory and internal state, particularly when inconsistencies arise.

% At the organizational level, the \textbf{organizational structure} governs how a collection of agents functions as a coherent system. Topology determines the pattern of connections, centralized, decentralized, hierarchical, or partitioned, that constrain how decisions propagate and where authority resides. Task and role differentiation, whether predefined, dynamically reassigned, or emergent through interaction, dictates the division of labor that underpins efficiency and adaptability. Finally, scalability and stability mechanisms determine whether the system can sustain large populations in practice and prevent the collective from collapsing into incoherence or disorder.

\subsection{Game Taxonomy for LLMGA Design}

The way a game agent is designed cannot be isolated from the environment in which it operates: Different game genres foreground distinct capabilities and place different challenges on agent design. For example, action games like Street Fighter demand far quicker reactions than strategy games like Poker, while requiring much less reasoning. Therefore, a taxonomy that captures how these characteristics shape agent design is essential.

Clarke et al.~\cite{clarke2017video} critically examine how conventional video game genre classifications often mix orthogonal dimensions such as mechanics and player structures, thereby lacking conceptual clarity. Building on this insight, we ground our taxonomy in established game studies literature through a gameplay-oriented perspective, drawing on the top-level groupings from SteamDB~\cite{steamdb2025} and the classification proposed by  et al.~\cite{lee2014facet}. To maintain coherence with existing LLMGA studies, we merge narrower categories (e.g., driving/racing, fighting) and additionally include sandbox games, resulting in six major genres as depicted in Table~\ref{tab:taxonomy}. Building on this categorization, we further introduce a challenge-centered view, where each genre is linked to the core design challenge that most strongly drives agent development. 

As shown in Table~\ref{tab:taxonomy}, we identify six representative game genres, each posing distinct design challenges for LLM-based agents. (1) Action games~\cite{bellemare2013arcade,llm-colosseum} unfold in real time and emphasize reflexive control, such as aiming, dodging, or chaining combos under tight temporal constraints. The core challenge is low-latency response, which shapes agent design by requiring fast action and hybrid architectures that reconcile LLM reasoning with frame-level responsiveness; (2) Adventure games~\cite{wang2022scienceworld,hausknecht2020interactive} emphasize exploration and long-horizon quests, where progress depends on remembering locations, items, and unresolved preconditions. The challenge is stateful world modeling, pushing agents to develop memory structures that maintain coherent records of evolving environments and dependencies; (3) Role-playing games~\cite{werewolf_rl,diplomacy} center on character customization, where players assume predefined roles with distinct traits and narrative trajectories~\cite{klevjer2012enter}. The key challenge is role fidelity, shaping agent design toward embedding role profiles into memory and reasoning so that dialogue and actions remain persona-consistent over extended horizons; (4) Strategy games~\cite{hu2025pokellmon,starcraftii} involve multi-step planning against adaptive adversaries, ranging from fully observable board games to imperfect-information settings with hidden states. Their central challenge is opponent-aware planning, which requires agents to integrate multi-step reasoning with theory-of-mind style opponent modeling; (5) Simulation games~\cite{park2023generative} approximate real-world or systemic processes, from individual social life to the evolution of societies. The challenge is dynamics fidelity, shaping agent design to ensure that behaviors remain credible and human-like rather than drifting into unrealistic patterns; (6) Sandbox games~\cite{minecraft,hafner2021benchmarking} offer open-ended environments where players set their own objectives, explore, and build. The challenge is open-ended goal progression, which drives designs where agents can generate self-directed goals, decompose them hierarchically, and accumulate reusable skills to sustain long-term play.

\begin{table}[t]
\centering
\caption{Gameplay taxonomy: game genres, core challenges, and representative environments.}
\vspace{-0.2cm}
\footnotesize
\setlength{\tabcolsep}{6pt}
\renewcommand{\arraystretch}{1.12}
\rowcolors{2}{gray!6}{white} % 如需纯黑白，注释掉
\begin{tabularx}{\textwidth}{@{} l l Y @{}}
\toprule
\textbf{Genre} & \textbf{Core Challenge} & \textbf{Representative Environments} \\
\midrule
\includegraphics[height=1.2em]{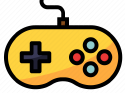} Action games & Low-latency response &
\makecell[l]{Atari 2600 games~\mbox{\cite{bellemare2013arcade}}; Procgen~\mbox{\cite{cobbe2020leveraging}};
ViZDoom~\mbox{\cite{kempka2016vizdoom}}; \\ 
DeepMind Lab~\mbox{\cite{beattie2016deepmind}}; 
Street Fighter~\mbox{\cite{llm-colosseum}}} \\
\includegraphics[height=1.2em]{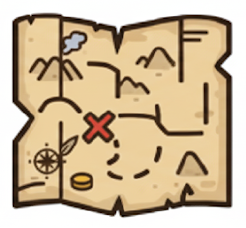} Adventure games & Stateful world modeling &
\makecell[l]{TextWorld~\mbox{\cite{cote2019textworld}}; Jericho~\mbox{\cite{hausknecht2020interactive}}; ALFWorld~\mbox{\cite{shridhar2021alfworld}};\\
ScienceWorld~\mbox{\cite{wang2022scienceworld}}; Red Dead Redemption II~\mbox{\cite{tan2024towards}} \\ \add{STARLING~\mbox{\cite{basavatia2024starling}}}} \\
\includegraphics[height=1.2em]{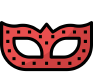} Role-playing games & Role fidelity &
\makecell[l]{AvalonBench~\mbox{\cite{light2023avalonbench}}; Werewolf~\mbox{\cite{wereworlf1}}; Diplomacy~\mbox{\cite{diplomacy}};\\
 \add{Among Us~\mbox{\cite{milkowski2026amongus}}; SOTOPIA~\mbox{\cite{zhou2023sotopia}}; clembench~\mbox{\cite{chalamalasetti2023clembench}}}} \\
\includegraphics[height=1.2em]{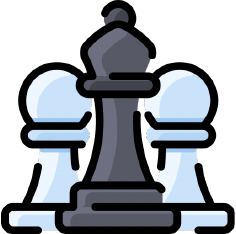} Strategy games & Opponent-aware planning &
\makecell[l]{Chess/Go~\mbox{\cite{feng2024chessgpt,toshniwal2022chess}}; Poker~\mbox{\cite{GoodPoker,huang2024pokergpt}};\\
Pokémon Battles~\mbox{\cite{hu2024pokellmon}}; StarCraft II~\mbox{\cite{starcraftii}} \\ \add{Card Games~\mbox{\cite{wang2025cardgames}}; LLM-PySC2~\mbox{\cite{li2025pysc2}}}} \\
\includegraphics[height=1.2em]{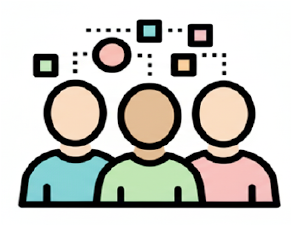} Simulation games & Dynamics fidelity &
\makecell[l]{Generative Agents~\mbox{\cite{park2023generative}}; Humanoid Agents~\mbox{\cite{wang2023humanoid}}; \\AgentSims~\mbox{\cite{lin2023agentsims}};
LyfeGame~\mbox{\cite{kaiya2023lyfe}}; CivRealm~\mbox{\cite{qi2024civrealm}}; \\
Artificial Leviathan~\mbox{\cite{leviathan}} \\ \add{IndoorWorld~\mbox{\cite{wu2025indoorworld}}; Moltbook~\mbox{\cite{li2026moltbook}}}} \\
\includegraphics[height=1.2em]{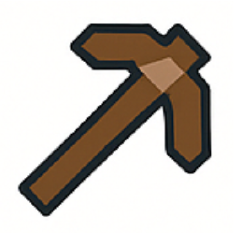} Sandbox games & Open-ended goal progression &
\makecell[l] {Minecraft~\mbox{\cite{minecraft}}; MineDojo~\mbox{\cite{fan2022minedojo}}; Crafter~\mbox{\cite{hafner2021benchmarking}} \\ \add{Plancraft~\mbox{\cite{dagan2025plancraft}}; UnrealZoo~\mbox{\cite{zhong2025unrealzoo}}}} \\
\bottomrule
\end{tabularx}
\label{tab:taxonomy}
\vspace{-0.3cm}
\end{table}

\section{Memory System of LLMGA}
\label{sec:memory}

LLMGAs require memory systems that encode and retain prior experience to ensure coherent and efficient interaction. Following classic distinctions in cognitive psychology~\cite{BaddeleyHitch1974,baddeley2012working}, we conceptualize an agent’s memory as working memory and long-term memory. 

In cognitive psychology, working memory functions as a transient and limited-capacity buffer that temporarily stores and manipulates information needed for ongoing cognitive processing~\cite{BaddeleyHitch1974,baddeley2012working}. In LLMGAs, this role is fulfilled by the model’s short context window and auxiliary mechanisms that keep recent observations ``in mind''. For working memory, we examine three key mechanisms. The first is \textbf{context extension}, which enlarges the effective context window so that recent events can be accommodated within short-term processing. The second is \textbf{memory compression}, which condenses lengthy inputs into compact representations, reducing capacity limits while preserving essential content. The third is \textbf{active maintenance}, which explicitly preserves recent bindings, plans, and intermediate states, preventing short-term drift and inconsistency caused by temporal decay.  

\begin{figure*}[htbp]
\centering
% \vspace{-0.1cm}
\includegraphics[width=14cm]{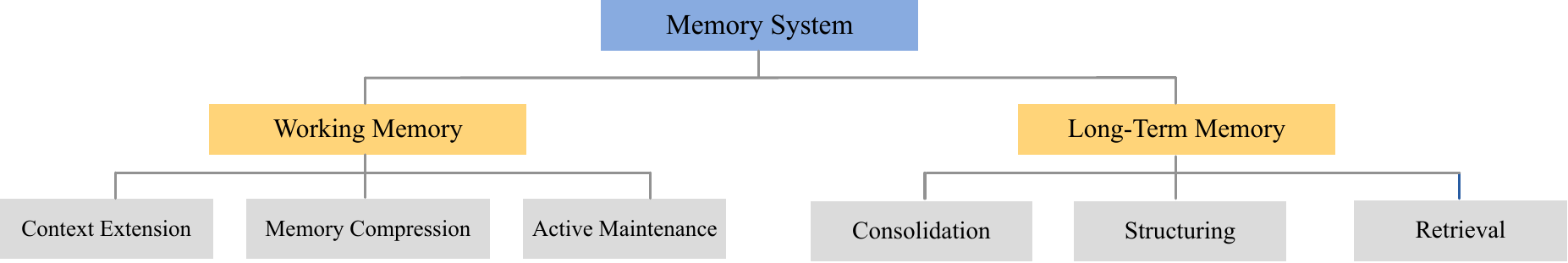}
\vspace{-0.5cm}
\caption{Overview of the memory system of LLMGAs.}
% \vspace{-0.2cm}
\vspace{-0.3cm}
\label{fig:memory_overview}
\end{figure*}

In contrast, long-term memory refers to the durable store of information that persists over extended periods beyond the limited span of working memory~\cite{tulving1972episodic,squire2004memory}. An LLMGA operating over long horizons faces three fundamental challenges naturally. As the storage is limited and most observations are transient, noisy, or low-value, the agent needs mechanisms to decide not only what information should persist beyond the immediate context, but also when transient traces should be committed to durable storage, so that memory remains useful and tractable. This motivates the role of memory \textbf{consolidation}. Once memories are retained, raw observations alone are insufficient to support abstraction, generalization, or efficient access. This creates the need to organize experiences into structured representations, motivating memory \textbf{structuring}. Finally, because decision-making in a given situation typically depends on only a small subset of relevant past memories, the agent needs to selectively reactivate critical information into working memory, motivating memory \textbf{retrieval}. Figure~\ref{fig:memory_overview} presents the structure of this section of different components within the memory system.

% In contrast, long-term memory refers to the durable store of information that persists over extended periods and can be retrieved to guide future behavior~\cite{tulving1972episodic,squire2004memory}. 
% It enables the accumulation of experience and knowledge that extend beyond the limited span of working memory~
% \cite{squire2004memory}. In LLMGAs, long-term memory is primarily realized through external storage systems that persist across interactions, such as vector databases, knowledge graphs, or serialized logs that record and retrieve past experience~\cite{park2023generative}. In addition, long-term memory can also be embedded within the parameters of the model itself, encoding generalized knowledge and experience that can be implicitly retrieved during generation~\cite{shao2023character}. 

% For long-term memory, we introduce mechanisms that enable agents to persist and exploit information across episodes. The first is memory \textbf{consolidation}, which decides when and what to commit from working memory to durable storage. The second is memory \textbf{structuring}, which determines how stored content is organized to facilitate abstraction and efficient access. The third is memory \textbf{retrieval}, which reactivates relevant past knowledge so that prior experience can inform ongoing decision-making. Figure~\ref{fig:memory_overview} presents the structure of this section of different components within the memory system.

% overview of the memory system.

\subsection{Working Memory}

Recent studies can be grouped into three categories. First, capacity extension enlarges the effective span of working memory by expanding positional encodings or restructuring attention. Second, memory compression distills lengthy or redundant input into more salient representations, mirroring the cognitive process of recoding multiple stimuli into higher-order units to overcome capacity limits~\cite{cowan2001magical}. Finally, active maintenance explicitly preserves variable bindings and states over short time scales, mirroring the human use of rehearsal to prevent rapid forgetting and inconsistency due to temporal decay~\cite{BaddeleyHitch1974,cowan2001magical}.

\textbf{Context Extension.} Context refers to the input tokens that the LLM can access when generating a new token, which is bounded by its context length~\cite{GPT3}. To overcome this, recent research focuses on extending the effective scope of the context window without full retraining. In LLM, position refers to the relative order of tokens within this context, typically represented through positional encodings that allow the model to distinguish token order in a sequence~\cite{transformer}. Position-based approaches modify positional encodings to enable length extrapolation.

Position Interpolation (PI) rescales Rotary Position Embeddings (RoPE) to support longer sequences with minimal fine-tuning. YaRN observes that uniform scaling distorts high-frequency positional components and instead applies non-uniform, frequency-aware 
interpolation that better preserves local token 
relationships~\cite{peng2023yarn}. LongRoPE further identifies that different RoPE dimensions tolerate different amounts of extension and searches for per-dimension rescaling factors, scaling context 
to over 2M tokens~\cite{ding2024longrope}. Beyond positional scaling, several methods restructure attention over long inputs. For example, Parallel Context Windows (PCW) processes long sequences by dividing them into coordinated segments without retraining~\cite{ratner2022parallel}, while PoSE enables generalization to longer contexts via sparse positional encoding~\cite{PoSE}.

\textbf{Memory Compression.} As input length grows, LLM performance often degrades due to limited capacity to maintain and manipulate multiple information items simultaneously~\cite{gong2024working}. To alleviate this bottleneck, memory compression techniques aim to condense long contexts into compact representations that preserve salient information. One line of work introduces soft or virtual tokens that summarize longer text spans, allowing the model to condition on compressed representations rather than the full input~\cite{lester2021power,ge2023incontext,gist}. Representative examples include AutoCompressor, which learns segment-level summary vectors~\cite{chevalier2023adapting}, and GIST, which trains virtual tokens to encode the essential content of a prompt for reuse~\cite{gist}. 

Another direction focuses on hierarchical summarization, where long sequences are recursively condensed into higher-level abstractions. Methods such as chain-of-summarization~\cite{starcraftii} incrementally condense game-state trajectories by segmenting the temporal sequence into short windows and recursively summarizing them into higher-level representations. Similarly, NUGGET~\cite{qin2023nugget} organizes long contexts into structured summaries, enabling efficient retrieval and reasoning while keeping inputs within the context window.

\begin{figure}[tbp]
    \centering
    \includegraphics[width=14cm]{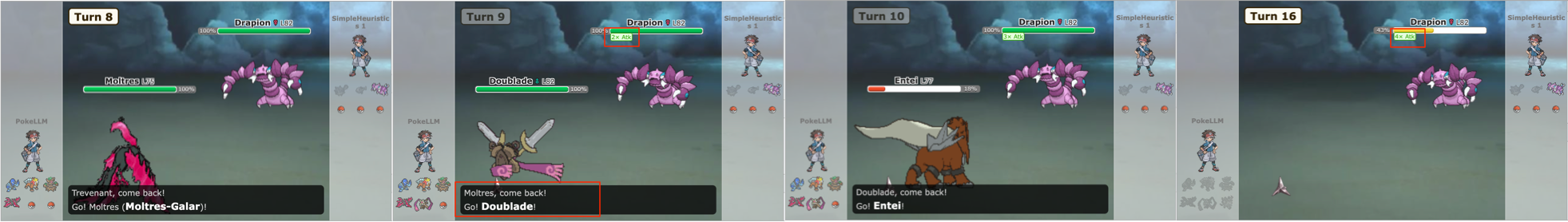}
    \vspace{-0.6cm}
    \caption{Illustration of temporal inconsistency~\cite{hu2025pokellmon}: When facing a powerful opponent, the agent tends to switch different Pokémon in consecutive steps rather than taking attack. }
    % \vspace{0.3cm}
    \vspace{-0.5cm}
    \label{fig:panic_switch}
\end{figure}

\textbf{Active Maintenance.} Active maintenance refers to preserving the contents of working memory over short intervals to ensure continuity in reasoning and action. In LLMGAs, failure to maintain such short-term state often leads to inconsistent decisions, even when recent events remain within the context window. A representative example is shown in PokéLLMon~\cite{hu2025pokellmon} (Figure~\ref{fig:panic_switch}), where agents repeatedly switch Pokémon in consecutive turns instead of attacking, and the issue is further exacerbated when chain-of-thought reasoning is adopted.

From the perspective of generation, reasoning introduces cumulative stochasticity that can lead to divergent decisions. Self-Consistency CoT (SC-CoT)~\cite{SC} attempts to mitigate this inconsistency by applying majority voting across reasoning paths in every step. One effective solution is Last-Thoughts~\cite{hu2025pokellmon}, which explicitly carries the reasoning trace from the previous step into the next prompt, anchoring decisions to prior deliberation and substantially improving consistency. Related approaches maintain short-term state by explicitly carrying compact summaries across steps. 
Belief-state maintenance summarizes the agent’s current understanding for reuse~\cite{li2023theory}, 
while MEM1~\cite{zhoumem1} and HiAgent~\cite{hu2024hiagent} update concise shared or subgoal-level memory states to retain salient information and improve short-horizon consistency. \add{In a similar spirit, StateFlow conceptualizes an agent's task-solving process as a state machine, explicitly tracking task state through states and transitions while delegating sub-tasks to actions~\cite{wu2024stateflow}.}

\subsection{Long-Term Memory}

% From a cognitive science perspective, long-term memory can be classified into three categories based on the content: 
% episodic memory that records specific experiences and events, semantic memory that retains factual and conceptual knowledge independent of context, and procedural memory that preserves skills and routines for action~\cite{tulving1972episodic,squire2004memory}. 

Recent agent architectures emphasize three fundamental processes in the design of long-term memory systems. First, memory consolidation determines when and what to commit from transient buffers to durable storage. Second, memory structuring addresses how stored content is organized. Finally, memory retrieval specifies how past knowledge is re-activated to guide ongoing decision-making. These components together ensure that long-term memory effectively archives past experience and supports future behavior.

\textbf{Consolidation.} In cognitive psychology, the transfer of information from working memory into long-term memory is termed consolidation~\cite{baddeley2012working,squire2004memory}. 
For LLMGAs, the analogous process is to decide when and what to commit from transient buffers to durable storage so that memory remains useful and tractable. A common paradigm is signal-triggered consolidation, where specific signals determine whether new information should be committed. In Generative Agents~\cite{park2023generative}, each incoming observation is assigned an importance score, and once the cumulative importance of recent events exceeds a threshold, the agent pauses to reflect, producing a summary that is then written into long-term memory. MemoryBank~\cite{memorybank} applies a similar principle, committing experiences when their relevance to the goal surpasses a salience threshold. Voyager~\cite{wang2023voyager} instead uses task outcomes as signals: successful code executions are committed into a skill library, while failed attempts are excluded or down-weighted.

More recent works extend write-back into more flexible learning-based schemes. For instance, CoALA~\cite{sumers2023cognitive} models “learning” as an explicit internal action within the agent’s action space, leaving it to the control policy (e.g., LLM) to decide when to encode new information into long-term memory. Self-Controlled Memory~\cite{wang2025enhancing} introduces a trainable memory controller that adaptively decides whether to write or use memory at each step. 
The controller is optimized jointly with the LLM through task-level supervision, such that memory updates are triggered only when they improve downstream performance. \add{Continual-learning agents extend this idea: NeSyC pairs an LLM with symbolic reasoning and a contrastive generality-improvement scheme that iteratively hypothesizes, validates, and refines reusable action knowledge, with a memory-based monitor triggering updates as embodied agents transition across ALFWorld, VirtualHome, and Minecraft~\cite{choi2025nesyc}. O3D mines large offline interaction logs to automatically discover reusable skills and distill generalizable knowledge, improving long-horizon decision-making~\cite{xiao2024o3d}.}

% Overall, write-back functions as the gateway between active experience and durable memory. For agent design, signal-triggered strategies provide a lightweight way to filter noisy environments, while policy-driven mechanisms allow for adaptive, context-sensitive consolidation. 
% By aligning write-back policies with task demands, memory systems can remain efficient while providing a foundation for richer agent behavior.  

% After deciding when to commit, an important design choice is how the memory is \textit{structured}. 

\begin{figure*}[tbp]
\centering
\includegraphics[width=14cm]{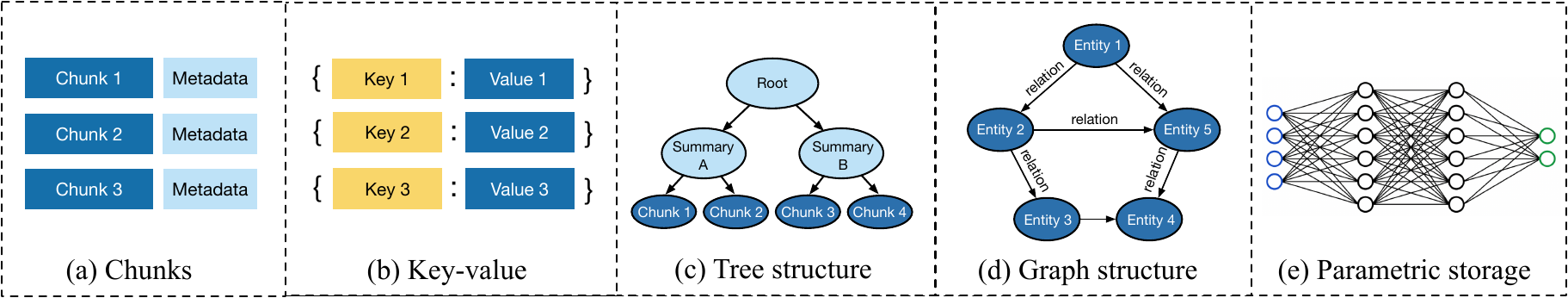}
\vspace{-0.6cm}
\caption{Illustration of representative memory structuring approaches.}
\vspace{-0.4cm}
\label{fig:memory_structure}
\end{figure*}

\textbf{Structuring.} Representative structures range from simple text fragments to highly organized graphs and implicit parametric storage, as shown in Figure~\ref{fig:memory_structure}. The most direct approach is to store observations as \textit{chunks}~\cite{park2023generative}. To facilitate later retrieval, each chunk can be augmented with metadata such as timestamps, importance scores~\cite{park2023generative}, or Q-values~\cite{zhang2024large}. \add{For example, MrSteve specializes such episodic chunks into a Place Event Memory that records what, where, and when each event occurred, enabling selective recall during Minecraft play~\cite{park2025mrsteve}.} Beyond raw fragments, many systems adopt a \textit{key–value} representation, where keys encode identifiers or semantic descriptors, and values store the corresponding content. This allows fast lookups and supports multimodal inputs: for example, Voyager represents keys as program descriptions paired with code snippets as values~\cite{wang2023voyager}, while JARVIS-1 stores visual observations as keys and successful execution plans as values~\cite{wang2023jarvis}. 

To capture hierarchical relations, memories can be recursively clustered into a \textit{tree} structure. Generative Agents~\cite{park2023generative}, RAPTOR~\cite{sarthi2024raptor}, and MemTree~\cite{memTree} all build memory trees where raw chunks form the leaves, and higher layers summarize increasingly abstract topics. Although the update mechanism differs, the underlying idea is to let new experiences traverse the tree, merging with existing nodes or forming new branches, while recursively updating parent summaries. \add{Game agents adopt similar hierarchies: GLoW maintains a dual-scale world memory pairing a global trajectory frontier of high-value discoveries with local multi-path reflection for hard-exploration text-adventure games~\cite{kim2026dualscale}, while MACLA distills trajectories into an external hierarchical procedural memory whose entries are tracked by Bayesian reliability estimates and contrastively refined~\cite{forouzandeh2025macla}.} An alternative design is to use \textit{graph}-structured memory. In knowledge graph approaches, nodes correspond to entities and edges correspond to semantic relations, typically extracted as triplets from text chunks, emphasizing fact representation~\cite{li2024graphreader,anokhin2024arigraph}. A-MeM~\cite{xu2025mem} organizes memory into a network of atomic notes enriched with tags and context, and edges represent semantic links between related notes, emphasizing interlinked note-taking and allowing updates to existing nodes. \add{KLPEG structures game elements, task dependencies, and causal relations into such a graph for multi-hop reasoning over update logs~\cite{mu2025klpeg}, and Optimus-1 pairs a hierarchical directed knowledge graph of world knowledge with a multimodal experience pool for long-horizon Minecraft tasks~\cite{li2024optimus}.}

Finally, some work explores \textit{parametric storage}, where memory is encoded implicitly in the model’s parameters rather than as external data. This perspective aligns with human cognition, which does not store verbatim text but instead internalizes experience. Fine-tuning on domain knowledge or episodic data can thus endow LLMs with embedded semantic or procedural memory~\cite{RedPajama2023,llama_rider}. For instance, CharacterLLM fine-tuned on synthetic character experiences can recall detailed knowledge of people, events, and objects in a role-consistent manner~\cite{shao2023character}.

% \vspace{0.1cm}
\textbf{Retrieval.} Memory retrieval is the process of reactivating stored information to guide current reasoning and action. Human studies also highlight that retrieval is selective and subject to recency, salience, and interference effects~\cite{Ebbinghaus1913,cowan2001magical}, which resonate with the design of LLMGAs. One common strategy is metadata-based retrieval, where memories are annotated with attributes such as timestamps, importance scores, or Q-values and ranked accordingly during retrieval~\cite{park2023generative,memorybank,zhang2024large}.

A second approach is semantic retrieval, where queries are embedded into a vector space and compared with stored representations. Generative Agents, for instance, compute cosine similarity between a self-instructed query and stored text memories~\cite{park2023generative}. In key–value settings, similarity is measured between the query and the key, with the associated value returned. This design allows flexibility across modalities: Voyager retrieves executable code by comparing program descriptions~\cite{wang2023voyager}, while JARVIS-1 retrieves action plans from multimodal keys that combine task descriptions and visual observations~\cite{wang2023jarvis}. \add{Building on example-based retrieval, agents can curate a database of their own successful trajectories and retrieve them as in-context exemplars, improving sequential decision-making on the ALFWorld and Wordcraft games without manual prompt engineering~\cite{sarukkai2025self}.} 

For more structured memories, retrieval can exploit graph or tree topologies. Graph-based retrieval begins by identifying relevant nodes using semantic or lexical cues, then traverses edges to explore multi-hop neighborhoods, finally synthesizing the resulting subgraph into a coherent narrative for the LLM to consume~\cite{li2024graphreader,anokhin2024arigraph}. \add{For Minecraft, goal-oriented graphs apply this graph traversal to objectives, so that retrieving a goal recursively surfaces its prerequisites for coherent multi-step plans~\cite{leung2025gogs}.} Tree-based retrieval instead performs hierarchical traversal: starting from the root, the agent selects top-$k$ relevant nodes at each level based on similarity, gradually descending to finer-grained leaves. Some variants collapse the hierarchy into a flat pool of summaries and retrieve based purely on semantic similarity~\cite{sarthi2024raptor,memTree}.  

Finally, for \textit{parametric storage}, knowledge is embedded implicitly in model weights rather than explicit structures. Such retrieval resembles implicit or procedural memory in humans, in which skills and habits are expressed without deliberate recall~\cite{shao2023character}. Table~\ref{tab:agent_memeory} summarizes representative LLMGAs by their memory design, showing the diversity of memory mechanisms across different game environments.

\begin{table}[htbp]
\centering
\small
\rowcolors{2}{gray!6}{white}
\vspace{-0.2cm}
\caption{Summary of representative LLMGAs in terms of memory design.}
\vspace{-0.2cm}
\resizebox{\textwidth}{!}{%
\begin{tabular}{@{}ccll@{}}
\toprule
\textbf{LLMGA} & \textbf{Environment} & \textbf{Working Memory} & \textbf{Long-Term Memory} \\ 
\toprule
Reflexion~\cite{shinn2023reflexion} & ALFWorld & In-episode experience & Reflection on previous episodes \\
% Xu et al.~\cite{wereworlf1} & Werewolf & In-episode experience & Reflective experience for retrieval \\ 
PokéLLMon~\cite{hu2025pokellmon} & Pokémon Battles & Active maintenance (last-step thoughts) & External game knowledge for retrieval \\
TextStarCraft~\cite{starcraftii} & StarCraft II & Memory compression (chain-of-summarization) &  \\
SuspicionAgent~\cite{guo2023suspicion} & Leduc Hold’em  & In-episode experience & Reflection on previous episodes\\ 
ProAgent~\cite{zhang2023proagent} & Overcooked-AI & Active Maintenance (Intention and belief) & Past experience for retrieval\\ 
% HAC~\cite{zhao2024hierarchical} & Minecraft & xxx & Successful multimodal plan for retrieval \\
Voyager~\cite{wang2023voyager} & Minecraft & Short-term code feedback & Successful code for retrieval \\
GTIM~\cite{zhu2023ghost} & Minecraft & Short-term action feedback  & Successful plan for retrieval \\ 
JARVIS-1~\cite{wang2023jarvis} & Minecraft & Short-term situational context & Successful multimodal plan for retrieval \\ 
GenerativeAgents~\cite{park2023generative} & Small Village & Memory compression (tree-based reflection) & Streaming memory with metadata \\
E2WM~\cite{xiang2024language} & VirtualHome & In-context dialogue & Exploration experience for fine-tuning\\
LLMPlanner~\cite{llm-planner} & ALFRED  & In-episode experience & Exemplar plan for retrieval \\ 
CharacterLLM~\cite{shao2023character} & Role-playing QA & In-context dialogue & Synthetic experience for fine-tuning\\
\add{Optimus-1~\cite{li2024optimus}} & \add{Minecraft} & \add{Short-term observation} & \add{Hybrid knowledge graph + experience pool} \\
\add{MrSteve~\cite{park2025mrsteve}} & \add{Minecraft} & \add{In-episode observation} & \add{Place Event Memory (what-where-when)} \\
\add{GLoW~\cite{kim2026dualscale}} & \add{Jericho} & \add{Local multi-path reflection} & \add{Dual-scale world memory} \\
\add{MACLA~\cite{forouzandeh2025macla}} & \add{ALFWorld} & \add{In-episode experience} & \add{Hierarchical procedural memory} \\
\add{KLPEG~\cite{mu2025klpeg}} & \add{Overcooked, Minecraft} &  & \add{Knowledge-graph memory} \\
\add{GoG~\cite{leung2025gogs}} & \add{Minecraft} &  & \add{Goal-oriented graph for retrieval} \\

\bottomrule
\end{tabular}}
\label{tab:agent_memeory}
\vspace{-0.2cm}
\end{table}

% \vspace{-0.2cm}
\section{Reasoning of LLMGA}
\label{sec:reasoning}

In cognitive science, reasoning is understood as the process of constructing and manipulating internal representations of known information to uncover implicit relations and abstract structures, thereby enabling conclusions that extend beyond what is explicitly given~\cite{johnson2010mental,evans2008dual}. In LLMGAs, reasoning serves as the central mechanism that transforms perceived and retrieved information into coherent plans, decisions, and explanations. It unfolds through language, by generating intermediate thought sequences that externalize internal deliberation and guide subsequent actions~\cite{CoT,kojima2022large}.

For instruction-guided reasoning, designed prompts elicit reasoning behavior directly at inference time. The first mechanism is \textbf{chain-of-thought}, which guides the model to articulate intermediate steps before arriving at an answer. The second is \textbf{search-based reasoning}, which explores multiple reasoning paths in parallel and selects among them to ensure consistency. The third is \textbf{reflective reasoning}, which iteratively improves reasoning across steps by incorporating internal self-critique or external signals. 

\begin{figure*}[htbp]
\centering
\includegraphics[width=14cm]{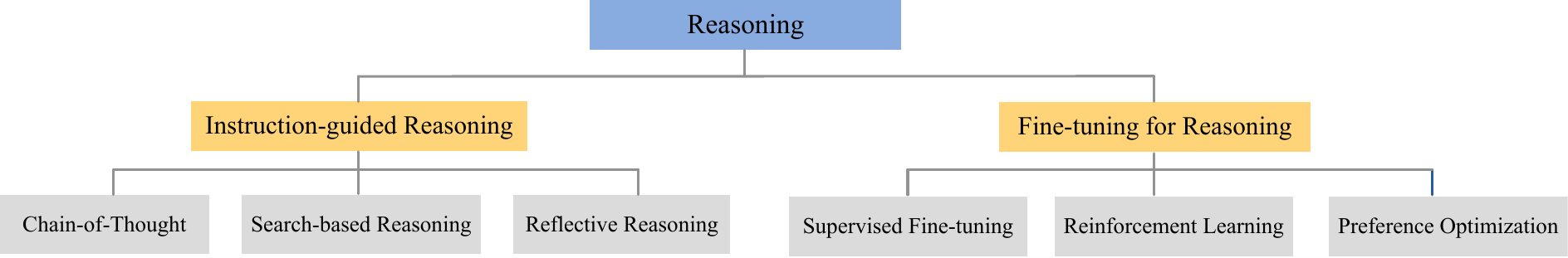}
\vspace{-0.6cm}
\caption{Categorization of reasoning mechanisms of LLMGAs.}
\vspace{-0.3cm}
\label{fig:reasoning_overview}
\end{figure*}

For fine-tuning paradigms, reasoning abilities are improved through optimization on data or experience interacted with the game environments. The first mechanism is \textbf{supervised fine-tuning}, where agents imitate expert demonstrations to acquire reasoning behaviors. The second is \textbf{reinforcement learning}, which updates policies or value models to optimize reasoning with task rewards. The third is \textbf{preference optimization}, which contrasts preferred and dispreferred generations to bias reasoning toward desirable outcomes. Figure~\ref{fig:reasoning_overview} presents the structure of this section of different components within the reasoning mechanism.

\subsection{Instruction-Guided Reasoning}

Prior studies have demonstrated that reasoning abilities can be elicited and amplified by deliberate prompting strategies at inference time, which guide models to externalize intermediate steps rather than relying solely on direct answer generation~\cite{CoT,kojima2022large}. We categorize existing methods into three groups. Chain-of-thought prompting elicits a single linear reasoning path, but is prone to error propagation. Search-based reasoning mitigates this by generating and organizing multiple trajectories to enhance robustness. Reflective reasoning emphasizes temporal refinement, where reasoning is iteratively improved using signals from prior experience or the environment.

\begin{figure*}[tbp]
\centering
\includegraphics[width=13.7cm]{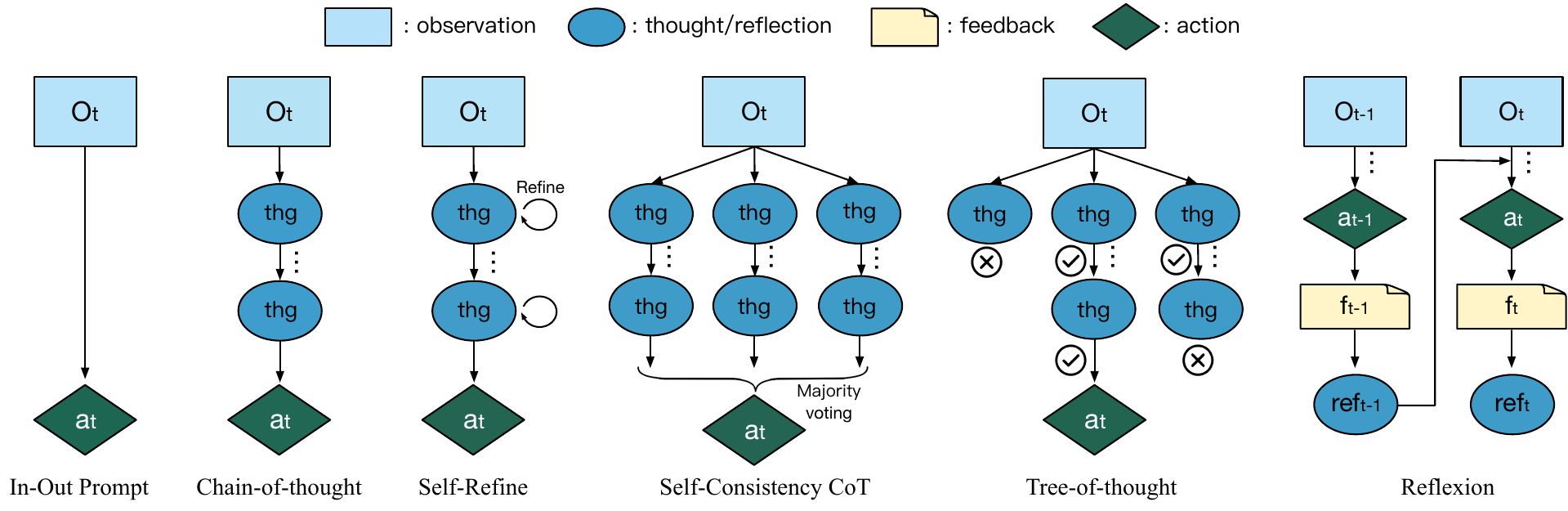}
\vspace{-0.2cm}
% \caption{Illustration of representative instruction-guided reasoning approaches.
% }
\caption{Illustration of representative instruction-guided reasoning approaches. $O_t$ denotes the observation at step $t$, $a_t$ denotes the action output at step $t$, and \textit{thg} denotes an intermediate reasoning step. In Reflexion, $f_t$ represents feedback and $ref_t$ denotes the resulting reflection at step $t$.}
\vspace{-0.3cm}
\label{fig:reasoning_approach}
\end{figure*}

\vspace{0.1cm}
\textbf{Chain-of-Thought.} CoT~\cite{CoT} is the basic approach that prompts LLMs to conduct intermediate reasoning before generating the answers, as shown in Figure~\ref{fig:reasoning_approach}. Since generation can be seen as an auto-regressive process of searching the next token in the latent space, the introduction of intermediate reasoning enhances the ability to traverse greater distances in that latent space, making LLMs capable of addressing more complex tasks. The ReAct~\cite{yao2022react} agent interleaves CoT reasoning and actions using few-shot prompting in text-based games. In their approach, reasoning acts as a mechanism for the agent to periodically check its task progress and plan its next steps. 

Intermediate reasoning introduces additional stochasticity, which can lead to inconsistent outputs. For instance, in Pokémon Battles, CoT may cause agents to panic-switch Pokémon in consecutive turns~\cite{hu2025pokellmon}, as shown in Figure~\ref{fig:panic_switch}. 
Moreover, once an early step deviates, subsequent tokens may inherit and magnify the error~\cite{madaan2024self}. Self-Refine~\cite{madaan2024self}, GPTLens~\cite{GPTLens} and RCI~\cite{computertask} aim to mitigate error propagation through self-criticism, first generating reasoning thoughts and then evaluating and refining them to improve the reasoning generation.

\vspace{0.1cm}
\textbf{Search-based Reasoning.} A major limitation of single-path chain-of-thought is fragility: randomness in sampling may yield inconsistent outputs, and early errors can propagate through the chain~\cite{SC,madaan2024self}. Search-based methods mitigate this by generating multiple intermediate reasoning candidates and then selecting, aggregating, or revising them. As shown in Figure~\ref{fig:reasoning_approach}, Self-Consistency~\cite{SC} alleviates inconsistency by prompting LLMs to generate multiple chains of thoughts independently, and conduct majority voting on the final answer to find the most consistent reasoning path. Tree-of-Thoughts~\cite{ToT} focuses on preventing error propagation by proposing multiple intermediate thoughts and selecting the correct one. Specifically, it decomposes a task into multiple steps, generates candidate thoughts for each step, and selects the most promising one, making the reasoning process resemble traversing a tree of thoughts. \add{THREAD frames generation as a thread of execution that dynamically spawns child threads for intermediate reasoning, improving adaptive decomposition~\cite{schroeder2024thread}.} For graph-based reasoning, Graph-of-Thoughts~\cite{besta2023graph} aggregates thoughts across different reasoning paths, converting a tree structure into a directed acyclic graph (DAG). SPRING~\cite{wu2024spring} constructs a template DAG in which each node corresponds to a question or instruction used to prompt LLMs for progressive reasoning. In their study, the authors prompt LLMs to summarize the Crafter paper~\cite{hafner2021benchmarking} into a DAG and then progressively traverse the DAG to answer these questions, thereby guiding the model through a step-by-step reasoning process. \add{AgentKit composes agent reasoning as a dynamic graph of modular natural-language nodes that support hierarchical planning and reflection in the Crafter sandbox game~\cite{wu2024agentkit}.}

\vspace{0.1cm}
\textbf{Reflective Reasoning.} Unlike generic LLM agents often evaluated on single-turn tasks, game agents operate within an observation–action–feedback loop, continuously perceiving the environment, taking actions, and adjusting decisions based on the resulting outcomes. Reflective reasoning builds on this loop by allowing agents to analyze the outcomes of their own actions and incorporate these reflections into future reasoning and behavior, as Reflexion~\cite{shinn2023reflexion} shown in Figure~\ref{fig:reasoning_approach}. This introduces a temporal dimension to reasoning, enabling the integration of experience over time.

Studies have shown that such temporal interaction enables LLMGAs to evolve over time by integrating feedback from past trajectories. The most direct form is reflection on failure: when an action fails, the agent can reuse the error signal to avoid repeating the same mistake. For instance, environments may provide explicit feedback such as “I cannot make a stone shovel because I need 2 more sticks” in MineCraft, which agents like Voyager~\cite{wang2023voyager} and GTIM~\cite{zhu2023ghost} exploit to iteratively refine their plans. Beyond explicit signals, reflective mechanisms such as Reflexion~\cite{shinn2023reflexion}, DEPS~\cite{wang2023describe}, and ProAgent~\cite{zhang2023proagent} guide agents to analyze their own chain-of-thought traces, identify where reasoning went wrong, and incorporate these insights into subsequent decisions. Even in environments with sparse feedback, agents can still benefit from heuristic signals~\cite{shinn2023reflexion}. \add{Recent game agents broaden the corrective signal beyond self-generated critiques. DGAP trains a discriminator from a few demonstrations to score how well each action aligns with optimal ones, then prompts the LLM to refine toward higher critic scores~\cite{qian2025dgap}. LEAP corrects a student agent with critiques from a teacher granted privileged state information only at training time, letting it surpass the teacher~\cite{choudhury2025leap}. Framing the LLM as an in-context reinforcement learner, scalar-reward feedback can be optimized over successive inference-time rounds~\cite{song2026icrl}. ReCAPA further mitigates cascading failures by predicting and contrasting deviations across actions, subgoals, and trajectories on benchmarks such as MineDojo~\cite{zeng2025recapa}.}

In addition to learning from failures, reflective reasoning can also benefit from reflecting on successes. Successful trajectories not only consolidate effective strategies but also provide contrastive signals when compared against failures. ExpeL~\cite{zhao2024expel} leverages this idea by retrieving the most relevant successful experiences, summarizing common patterns, and deriving insights through success–failure comparisons. Similarly, KWM~\cite{qiao2025agentplanningworldknowledge} extracts task knowledge from expert-demonstrated trajectories and distills it into a dedicated world knowledge model, which is then used to guide the agent’s planning in future episodes. \add{AutoManual has a builder agent refine interaction experience into human-readable rule manuals that curb hallucination, reaching high success on ALFWorld from a single demonstration~\cite{chen2024automanual}, while DiVE discovers, verifies, and evolves world-dynamics rules from a handful of demonstrations to reach human-level reward in Crafter and MiniHack~\cite{sun2024dive}.} In summary, reflective reasoning shares the basic idea of reinforcement learning that uses feedback to correct mistakes and reinforce successful strategies, embodying the principle of learning through interaction with the environment.

% Figure~\ref{fig:reasoning_approach} summarizes representative instruction-guided reasoning approaches introduced in this section.

\subsection{Fine-tuning for Improving Reasoning}

In this subsection, we examine fine-tuning techniques for optimizing reasoning and action generation. Based on the training strategy, existing methods can be grouped into three categories. Supervised fine-tuning learns from expert trajectories to imitate reasoning and action generation. Reinforcement learning updates policies with reward feedback, reinforcing reasoning and actions that lead to favorable outcomes. Preference optimization leverages comparisons between better and worse trajectories to align models without the need for explicit reward models. It is worth noting that some methods mentioned below optimize only the final action without explicit reasoning, however, they can be extended to improve reasoning by eliciting chain-of-thought, allowing reasoning to be shaped through its effect on action outcomes~\cite{Kahneman2011}.

\vspace{0.1cm}
\textbf{Supervised Fine-Tuning.} Supervised fine-tuning trains LLM agents on collected trajectories to maximize the likelihood of reproducing demonstrated reasoning and actions. The most common approach is behavior cloning, where agents directly imitate expert demonstrations. Such trajectories may come from human experts~\cite{reed2022a}, from state-of-the-art agents~\cite{lin2024swiftsage}, or from teacher LLMs that generate rollouts for training student models~\cite{zeng2023agenttuning}. Behavior cloning is widely adopted as an initialization strategy, providing a strong prior policy that can later be refined by reinforcement learning~\cite{yang2023octopus,ETO}.

Building on this idea, rejection sampling fine-tuning introduces a selection stage before training. Instead of imitating all trajectories, the model generates multiple candidates and filters them according to predefined criteria, such as binary success/failure signals or reward estimates. RFT~\cite{RFT}, for example, fine-tunes models only on successful trajectories, while other works employ environment-provided or model-estimated rewards to guide sample selection~\cite{touvron2023llama}. Although this improves data quality, it can be inefficient when the agent initially produces few successful rollouts. \add{Wang et al. show that supervised fine-tuning on high-quality gameplay data lets LLMs approach strong game AIs across eight complex card games and acquire proficiency in several games at once~\cite{wang2025cardgames}. ChessLLM supervised-fine-tunes a language model on complete chess games trajectories to reach a 1788 Elo against Stockfish, showing that training on longer game sequences yields a large rating advantage~\cite{zhang2025chessllm}. For socially intelligent agents, SOTOPIA-$\pi$ combines behavior cloning with self-reinforcement on social-interaction episodes filtered by LLM ratings, letting a 7B model match GPT-4's social-goal completion~\cite{wang2024sotopiapi}.}

\vspace{0.1cm}
\textbf{Reinforcement Learning.} Reinforcement learning (RL) provides another major paradigm for improving reasoning and action generation in LLM agents. Existing game agents~\cite{carta2023grounding,du2023guiding,TWOSOME,yang2023octopus} mainly adopt the Proximal Policy Optimization (PPO) algorithm~\cite{ppo}, where the model is trained as a policy $\pi(a_t \mid s_t)$ (without explicit reasoning) and updated using advantage-weighted gradients to favor actions leading to higher rewards. Alongside the policy model, PPO also learns a value function to estimate the relative quality of state–action pairs. While effective, applying RL to LLMs faces the challenge of an enormous generation space, which often leads to inadmissible actions. To address this, some methods compute the probability distribution of admissible actions by the chain rule before sampling, ensuring that the generated actions remain valid~\cite{carta2023grounding,TWOSOME}.

Recent works further integrate explicit reasoning into RL training, where the LLM is trained as a policy $\pi(rs_t, a_t \mid s_t)$. Reinforced Fine-Tuning (ReFT)~\cite{trung2024reft} introduces chain-of-thought supervision into PPO, encouraging the model to generate reasoning paths that lead to correct answers. However, because reasoning tokens are often much longer than action tokens, naive optimization can overweight reasoning relative to actions. Zhai et al.~\cite{zhai2025fine} propose downscaling the likelihood of reasoning steps, showing that moderate scaling achieves better balance between planning and acting. Beyond policy optimization, value-based methods such as Q-learning extend RL to reasoning by treating partial generations as states and token expansions as actions. This formulation allows the use of search algorithms, such as Best-of-N sampling or Monte Carlo tree search, to evaluate and expand reasoning paths guided by the Q-function~\cite{zhang2024rest}. \add{Game agents extend this search-integrated view: SEEA-R1 fuses Monte Carlo tree search with a Tree-GRPO objective and a multimodal generative reward model to self-evolve on long-horizon ALFWorld tasks~\cite{tian2025seea}.}

% More recently, DeepSeek-R1 and R1-Zero~\cite{guo2025deepseek} achieve remarkable problem-solving performance using Group Relative Policy Optimization (GRPO)~\cite{shao2024deepseekmath}, a PPO variant that eliminates the need for a value model by comparing reasoning paths within a group and using reward statistics as baselines.

A challenge is that conventional reward signals (and the value estimates derived from them) are provided only at the action level, providing no feedback on the intermediate reasoning steps. This causes error to propagate through the reasoning until the final outcome is known. To address this limitation, Process Reward Modeling (PRM)~\cite{prm} supplies dense feedback by explicitly evaluating intermediate reasoning steps. \add{In games, the reward signal and objective are themselves often tailored to the task: LARM derives rewards from a large LLM acting as a referee to train a compact policy for long-horizon Minecraft~\cite{li2025larm}, DVM imposes a win-rate-constrained decision-chain reward to make Werewolf agents controllable~\cite{zhang2025dvm}, GFlowVLM replaces scalar-reward maximization with reward-proportional generative flow networks for card games such as BlackJack~\cite{kang2025gflowvlm}, and DipLLM sets an approximate-equilibrium policy as the learning target for Diplomacy, matching Cicero with a fraction of the data~\cite{xu2025dipllm}. Two further paradigms sidestep hand-specified rewards. First, self-play and learning-through-play supply the signal directly: SPIRAL trains on zero-sum games such as TicTacToe and Kuhn Poker with role-conditioned advantage estimation~\cite{liu2026spiral}, SPAG plays both sides of Adversarial Taboo to improve general reasoning~\cite{cheng2024spag}, and an RL-instructed discussion policy is learned through One Night Ultimate Werewolf~\cite{jin2024onuw}. Second, LLMs shape the RL loop without acting as the policy: EnvGen generates and adapts training environments so a small agent surpasses GPT-4 agents on Crafter~\cite{zala2024envgen}, and language-guided exploration scores candidate decisions for an RL explorer in ScienceWorld~\cite{golchha2024lge}.}

% Although training PRMs often requires costly annotations or carefully engineered heuristics, recent work has explored scalable approximations. For example, MATH-SHEPHERD~\cite{wang2023math} estimates step-level rewards by Monte Carlo sampling of multiple continuations and averaging their outcomes, while FreePRM~\cite{yuan2024free} reduces annotation cost with self-supervised signals. 

\vspace{0.1cm}
\textbf{Preference Optimization.} The idea of preference optimization was first explored in games, where OpenAI demonstrated that human preference comparisons could be used to train reward models for Dota 2~\cite{christiano2017deep}. This principle of optimizing agents by favoring trajectories preferred by humans rather than relying on hand-crafted rewards later became the foundation for aligning language models. Building on this, Direct Preference Optimization (DPO)~\cite{DPO} enables contrastive training without an explicit reward model by maximizing the margin between preferred and non-preferred generations, thereby simplifying the optimization process and reducing cost. In the context of game agents, this preference-based framework can also be applied at the trajectory or step level: ETO~\cite{ETO} alternates between exploration and fine-tuning with DPO on successful vs. failed rollouts, while IPR~\cite{IPR} extends this to step-wise preference optimization, pairing reasoning steps according to the average reward calculated via Monte Carlo method. In Table~\ref{tab:agent_reasoning}, we list representative LLMGAs by their reasoning mechanism design, aligned with the two dimensions of our categorization.

\begin{table}[htbp]
\centering
% \small
\footnotesize
\rowcolors{2}{gray!6}{white}
\vspace{-0.5cm}
\caption{Summary of representative LLMGAs in terms of reasoning mechanism.}
\vspace{-0.3cm}
\resizebox{\textwidth}{!}{%
\begin{tabular}{@{}cclll@{}}
\toprule
\textbf{LLMGA} & \textbf{Environment} & \textbf{Instruction-guided Reasoning} & \textbf{Fine-tuning for Improving Reasoning} \\ 
\toprule
ReAct~\cite{yao2022react} & ALFWorld, \textit{etc.} & CoT &  \\ 
Reflexion~\cite{shinn2023reflexion} & ALFWorld, \textit{etc.} & CoT + Reflective reasoning &  \\
ADAPT~\cite{prasad2023adapt} & ALFWorld, \textit{etc.} & As-needed CoT (planning) &  \\ 
SwiftSAGE~\cite{lin2024swiftsage} & ScienceWorld & As-needed CoT (planning) &  \\ 
ETO~\cite{ETO} & ALFWorld, \textit{etc.} &  & Trajectory-level preference optimization \\
IPR~\cite{IPR} & ALFWorld, \textit{etc.} &  & Step-level preference optimization \\
GLAM~\cite{carta2023grounding} & BabyAI-Text &  & RL fine-tuning \\ 
TWOSOME~\cite{TWOSOME} & Overcooked-AI & & RL fine-tuning \\
Xu et al.~\cite{wereworlf1} & Werewolf & Reflective reasoning &  \\ 
Xu et al.~\cite{werewolf_rl} & Werewolf &  & RL-based candidate selection \\ 
WarAgent~\cite{worldwarii} & Diplomacy-like & Structural reasoning & \\ 
PokéLLMon~\cite{hu2024pokellmon,hu2025pokellmon} & Pokémon Battles & Consistent reasoning generation &  \\
ChessGPT~\cite{feng2024chessgpt} & Chess &  & Supervised fine-tuning \\ 
PokerGPT~\cite{huang2024pokergpt} & Texas Hold'em &  & RL from human feedback\\ 
SuspicionAgent~\cite{guo2023suspicion} & Leduc Hold’em & Theory-of-mind reasoning & \\ 
HLA~\cite{liu2023llm} & Overcooked & As-needed CoT (planning) &  \\ 
S-Agents~\cite{sagents} & Minecraft & Goal decomposition, evaluation & \\ 
HAC~\cite{zhao2024hierarchical} & Minecraft & Goal decomposition, correction, evaluation & \\
Voyager~\cite{wang2023voyager} & Minecraft & Code as policy, correction & \\
DEPS~\cite{wang2023describe} & Minecraft & Goal decomposition, reflection, selection & \\
GTIM~\cite{zhu2023ghost} & Minecraft & Goal decomposition, correction & \\ 
JARVIS-1~\cite{wang2023jarvis} & Minecraft & Goal decomposition, reflection &  \\ 
Plan4MC~\cite{yuan2023plan4mc} & Minecraft & Goal decomposition & \\
RL-GPT~\cite{rlgpt} & Minecraft & Reasoning as code generation &  \\ 
LLaMARider~\cite{llama_rider} & Minecraft &  & Novelty-driven Supervised fine-tuning \\ 
Project Sid~\cite{sid} & Minecraft & Social awareness reasoning &  \\
GenerativeAgents~\cite{park2023generative} & Sims-like game & Tree-based reflection \& planning & \\
HumanoidAgents~\cite{wang2023humanoid} & Social & Affective-driven planning & \\ 
LLMPlanner~\cite{llm-planner} & ALFRED & Planning \& re-planning &  \\ 
Octopus~\cite{yang2023octopus} & OctoVerse & Reasoning as code generation & RL fine-tuning \\
ELLM~\cite{du2023guiding} & Crafter & Situated goal generation & \\ 
SPRING~\cite{wu2024spring} & Crafter & Structural reasoning & \\ 
% OMNI~\cite{zhang2024omni} & Crafter & - & Interesting goal selection \\
\add{PokéChamp~\cite{karten2025pokechamp}} & \add{Pokémon Battles} & \add{Minimax tree search} & \add{} \\
\add{Strategist~\cite{light2025strategist}} & \add{Avalon, GOPS} & \add{Bi-level tree search} & \add{} \\
\add{Schultz et al.~\cite{schultz2025boardgames}} & \add{Chess, Hex} & \add{External/internal search} & \add{} \\
\add{DiffuSearch~\cite{ye2025diffusearch}} & \add{Chess} & \add{Implicit diffusion search} & \add{} \\
\add{MC-DML~\cite{shi2025mcdml}} & \add{Jericho} & \add{MCTS with memory} & \add{} \\
\add{AutoManual~\cite{chen2024automanual}} & \add{ALFWorld} & \add{Reflective rule learning} & \add{} \\
\add{AgentKit~\cite{wu2024agentkit}} & \add{Crafter} & \add{Dynamic-graph reasoning} & \add{} \\
\add{DGAP~\cite{qian2025dgap}} & \add{ScienceWorld} & \add{Discriminator-guided refinement} & \add{} \\
\add{DiVE~\cite{sun2024dive}} & \add{Crafter} & \add{World-dynamics reflection} & \add{} \\
\add{ReCAPA~\cite{zeng2025recapa}} & \add{MineDojo} & \add{Hierarchical predictive correction} & \add{} \\
\add{DipLLM~\cite{xu2025dipllm}} & \add{Diplomacy} & \add{} & \add{RL fine-tuning} \\
\add{SPIRAL~\cite{liu2026spiral}} & \add{Zero-sum games} & \add{} & \add{Self-play RL} \\
\add{SPAG~\cite{cheng2024spag}} & \add{Adversarial Taboo} & \add{} & \add{Self-play RL} \\
\add{ChessLLM~\cite{zhang2025chessllm}} & \add{Chess} & \add{} & \add{Supervised fine-tuning} \\
\add{Card Games~\cite{wang2025cardgames}} & \add{Card games} & \add{} & \add{Supervised fine-tuning} \\
\add{LARM~\cite{li2025larm}} & \add{Minecraft} & \add{} & \add{Referee-based RL} \\
\add{SEEA-R1~\cite{tian2025seea}} & \add{ALFWorld} & \add{} & \add{Tree-structured RL} \\
\add{DVM~\cite{zhang2025dvm}} & \add{Werewolf} & \add{} & \add{RL fine-tuning} \\
\add{GFlowVLM~\cite{kang2025gflowvlm}} & \add{BlackJack} & \add{} & \add{GFlowNet fine-tuning} \\
\bottomrule
\end{tabular}}
\label{tab:agent_reasoning}
\vspace{-0.2cm}
\end{table}

\section{Perception and Action Interfaces of LLMGA}
\label{sec:in/output}

\begin{figure*}[htbp]
\centering
\includegraphics[width=14cm]{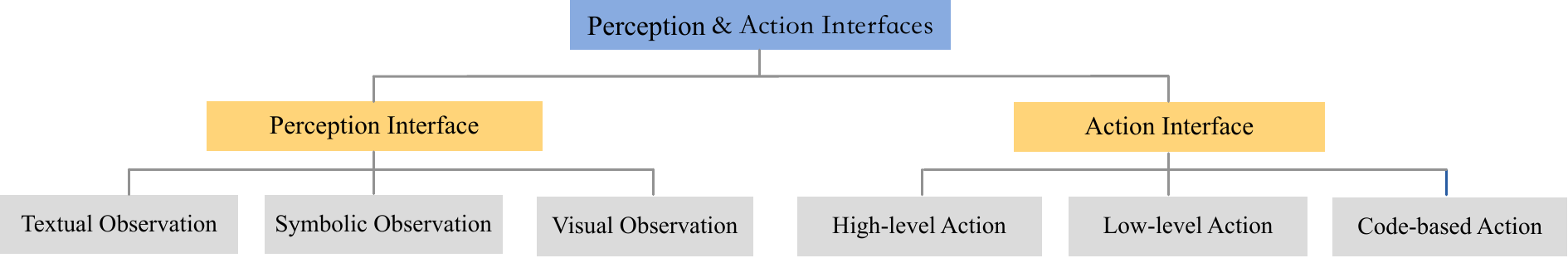}
% \vspace{-0.1cm}
\vspace{-0.5cm}
\caption{Overview of perception and action interfaces in LLMGAs.}
\vspace{-0.5cm}
\label{fig:interface_overview}
\end{figure*}

LLMGAs differ from generic LLM systems in that they operate within a continuous perception-action loop. To support this loop, agents rely on perception and action interfaces that serve as their eyes and hands for interacting with the environment~\cite{wang2023voyager,hu2024pokellmon}. On the input side, the perception interface determines how raw game states are abstracted into representations that can be processed by the LLM, handling \textbf{textual}, \textbf{symbolic}, and \textbf{visual} observations. On the output side, the action interface ensures that the model’s decisions are translated into admissible in-game operations by grounding the LLM outputs into \textbf{high-level}, \textbf{low-level}, and \textbf{code-based} actions. Figure~\ref{fig:interface_overview} outlines the structure of this section.

\subsection{Perception Interface}

The perception interface defines how an LLMGA accesses and processes information from the game environment. The most direct and widely adopted way to categorize input-processing methods is based on the modality of the game observation, such as textual, symbolic, or visual forms.

\textbf{Textual Observations.} In text-based or dialogue-centric games~\cite{zork1980,shridhar2021alfworld}, the environment state is natively presented in natural language. 
In such cases, the agent can directly consume text descriptions as observations without additional preprocessing~\cite{yao2022react,shinn2023reflexion}. 
This modality is straightforward, as it aligns with the input format of LLMs, but it is restricted to environments where language is the primary medium of interaction.

\vspace{0.1cm}
\textbf{Symbolic Observations.} Some video game environments provide structured state information through APIs or game engines~\cite{starcraftii,hu2024pokellmon,li2023cooperative,minecraft}. 
These symbolic variables (e.g., avatar health, inventory, world coordinates or object properties) can be transformed into a form that the LLM can process, often through textual summaries or structured prompt templates. 
For example, Mineflayer~\cite{mineflayer} exposes a Minecraft character’s stats and surrounding entities, which can then be summarized into a natural-language prompt~\cite{hu2024pokellmon}. 
Symbolic observations are efficient when the selected variables can sufficiently capture the essential context, but they risk losing fidelity in complex environments where subtle but critical distinctions, such as object textures, spatial relations, or small visual cues, are omitted from the symbolic representation.

\vspace{0.1cm}
\textbf{Visual Observations}. In video games, the agent typically perceives the environment as a sequence of rendered images. 
Since LLMs cannot directly operate on raw pixels, the perception interface requires a translator that converts visual signals into interpretable representations. 
One approach is \textit{vision-to-text translation}, where object detectors or pretrained encoders such as CLIP~\cite{CLIP} produce captions or object lists that can be inserted into prompt templates.  For example, an agent in a 3D environment can use an object detector to list visible objects (“a key on the floor, a locked door ahead”) and insert them into the prompt template~\cite{zhang2023building,llm-planner}. The agent can also adopt a visual encoder to map images into pre-defined text descriptions~\cite{wang2023describe,wang2023jarvis,du2023guiding}, or a text decoder to generate the caption~\cite{du2023guiding} to summarize the scene.

An alternative is to use \textit{multimodal LLMs} to directly process raw frames. These models align visual and textual information in a shared representation space, allowing an agent to feed raw images or pixels to the model and get an immediate understanding. Recent works~\cite{de2024will,tan2024towards} leverage general-purpose multimodal LLMs (e.g., GPT-4 Vision~\cite{GPT4}) to interpret game visuals. This direct approach can generalize well to new games, but still requires additional mechanisms to correct errors or uncertainties in its perceptions~\cite{yang2023octopus,tan2024towards}. Game-specific multimodal models have also been introduced, \textit{e.g.}, LLMs finetuned on paired image-instruction data for a particular game, such as SteveEye~\cite{SteveEye} or learned from
environmental feedback through RL such as Octopus~\cite{yang2023octopus}. \add{STEVE follows this direction in Minecraft, coupling a vision-perception module that interprets visual observations with an LLM for reasoning and a code-action skill database, trained on the STEVE-21K dataset of vision--environment, QA, and skill-code pairs~\cite{zhao2024seethink}. Rather than passively consuming observations, ActiveVOO performs value-of-observation guided active sensing for partially observable embodied planning, quantifying the utility of sensing actions from LLM and VLM commonsense priors so the agent perceives only task-relevant objects in ALFWorld~\cite{liu2025activevoo}.}

\subsection{Action Interface}

The action interface determines how an LLM-based agent’s decisions are grounded into executable operations within the game environment. 
Unlike generic LLM outputs that produce unconstrained text, games require actions that conform to specific control formats. Accordingly, action interfaces are categorized by the type of action required by games: \textit{high-level actions} represent semantic or logical operations (e.g., “open the door”); \textit{low-level actions} specify concrete control signals such as keystrokes or mouse movements; and \textit{programmatic actions} output structured commands or API calls that the environment can directly execute.

\textbf{High-Level Actions.} In games where actions are expressed as discrete choices~\cite{hu2025pokellmon}, the generation problem can be reformulated as a selection task. In this case, the model can simply select one of the provided options as the action. In parser-based environments, such as text adventure games or interactive narratives~\cite{TextWorldProblemsComp,hausknecht2020interactive}, the LLM must generate a command that follows specific syntax, such as  “open the door” or “pick up the sword”. Outputs that deviate from the expected syntax are treated as invalid and ignored. Therefore, the core challenge is to ensure that output actions are admissible. Recent work has introduced correction mechanisms, such as mapping generated phrases to the closest permissible action~\cite{huang2022language}. An alternative is constrained decoding: instead of unconstrained token-by-token decoding, it computes the joint likelihood of each valid action sequence using the chain rule, and then normalize across the entire action set~\cite{carta2023grounding}. However, such token-level probabilities penalize longer commands disproportionately, leading to systematic bias against valid but longer actions. To mitigate this problem, TWOSOME~\cite{TWOSOME} introduces length normalization by scaling log-likelihoods with the action’s token count, thereby balancing the probability distribution over admissible actions. \add{Other game agents refine high-level action selection with auxiliary knowledge or value signals: KnowAgent constrains executable-action planning with an explicit action knowledge base to reduce hallucination~\cite{zhu2025knowagent}, while reinforced advantage feedback (ReAd) trains a critic that estimates per-agent advantage values so agents can reject low-value actions without costly verification in the Overcooked-AI~\cite{zhang2024read}.}

\textbf{Low-Level Actions.} Low-level actions operate at the control layer, such as keystrokes, mouse movements, joystick inputs, and are executed at each timestep. A low-level controller (policy) is responsible for translating a high-level action from the LLM into a sequence of control signals. One approach is heuristic planning~\cite{multi_agent_coordination,liu2023llm,park2023generative}: given an intent such as “chop a tree,” the system invokes a path planner (e.g., A$^{*}$) to locate the nearest tree and then issues the necessary movement and interaction commands. Another approach is to learn a low-level controller (policy) that generates the required action sequences to realize the LLM’s high-level decisions. Such policies can be trained either through imitation learning from expert demonstrations or through reinforcement learning with environment feedback, often aided by goal-conditioned rewards or semantic similarity between goals and observed transitions~\cite{rlgpt}.

\textbf{Code-based Actions.} Code-based actions express agent decisions as structured code or API calls that can be executed directly in the environment~\cite{wang2023voyager,tan2024towards}. Their structured nature provides explicit semantics and eliminates ambiguity, allowing complex operations to be specified with precision (e.g., \textit{bot.equip(sword);} through a modding API~\cite{mineflayer} or \textit{key\_press("M")} at the system level). A further advantage is verifiability: code outputs can be parsed and checked before execution, and compilers or interpreters supply syntax feedback that enables automatic detection and correction of invalid commands~\cite{wang2023voyager}. In addition, programmatic actions support reusability by enabling agents to maintain a library of high-level primitives that encapsulate recurring skills. These functions can be flexibly composed, reducing redundant low-level generation and facilitating scalable, compositional behavior~\cite{tan2024towards}. \add{Recent work makes code-based grounding more learnable: LearnAct lets an agent analyze failed attempts and iteratively create and revise executable Python functions that expand its action space~\cite{zhao2024learnact}. CoPiC emits multiple planning programs and trains a domain-adaptive selector to pick the plan best aligned with long-term rewards on ALFWorld, NetHack, and StarCraft II unit building~\cite{tian2026copic}. MaestroMotif has an LLM specify per-skill reward functions and generate code that trains and composes RL-learned skills in the NetHack~\cite{klissarov2025maestromotif}.} Table~\ref{tab:agent_perception_action} lists representative LLMGAs, categorized by their perception and action interfaces.

\begin{table}[htbp]
\centering
% \small
\footnotesize
\rowcolors{2}{gray!6}{white}
\vspace{-0.1cm}
\caption{Summary of representative LLMGAs in terms of perception \& action interfaces.}
\vspace{-0.4cm}
\resizebox{\textwidth}{!}{%
\begin{tabular}{@{}ccllll@{}}
\toprule
\textbf{Agent} & \textbf{Game} & \textbf{Perception Interface} & \textbf{Action Interface} \\ 
\toprule
ReAct~\cite{yao2022react} & ALFWorld, \textit{etc.} & Textual input & High-level action & \\ 
SwiftSAGE~\cite{lin2024swiftsage} & ScienceWorld & Textual input & High-level action \\ 
% GLAM~\cite{carta2023grounding} & BabyAI-Text & - & Fine-tuned with RL \\ 
Cradle~\cite{tan2024towards} & RDR2 & Visual input (VLM) & Low-level action (via keyboard–mouse control APIs) \\ 
Xu et al.~\cite{werewolf_rl} & Werewolf & Textual input & High-level action \\ 
PokéLLMon~\cite{hu2024pokellmon} & Pokémon Battles & Symbolic input & High-level action \\
TextStarCraft~\cite{starcraftii} & StarCraft II & Symbolic input & Low-level action (rule-based controller) \\
ChessGPT~\cite{feng2024chessgpt} & Chess & Symbolic input & High-level action \\ 
% PokerGPT~\cite{huang2024pokergpt} & Texas Hold'em & Symbolic input  & High-level action\\ 
SuspicionAgent~\cite{guo2023suspicion} & Leduc Hold’em & Symbolic input & High-level action \\ 
ProAgent~\cite{zhang2023proagent} & Overcooked-AI & Symbolic input & Low-level action (via path search + API calls) \\ 
TWOSOME~\cite{TWOSOME} & Overcooked-AI & Symbolic input & High-level action (admissible action generation) \\ 
Voyager~\cite{wang2023voyager} & Minecraft & Symbolic input & Code-based action (via Mineflayer code execution) \\
GTIM~\cite{zhu2023ghost} & Minecraft & Symbolic input & Low-level action (via API calls) \\
JARVIS-1~\cite{wang2023jarvis} & Minecraft & Visual and symbolic input & Low-level action (via controller and API calls) \\ 
CoELA~\cite{zhang2023building} & TDW-T\&WAH & Visual input (object detector) & Low-level action (via rule-based controller) \\ 
GenerativeAgents~\cite{park2023generative} & Small Village & Textual input & High-level actions\\
ZeroShotPlanner~\cite{huang2022language} & VirtualHome & Symbolic input & High-level actions (semantic translation) \\
% Octopus~\cite{yang2023octopus} & OctoVerse & - & &  \\
ELLM~\cite{du2023guiding} & Crafter & Visual input (visual encoder) & Low-level action (RL-based controller)  \\
\add{STEVE~\cite{zhao2024seethink}} & \add{Minecraft} & \add{Visual input (perception module)} & \add{Code-based action (skill database)} \\
\add{LearnAct~\cite{zhao2024learnact}} & \add{ALFWorld} & \add{Textual input} & \add{Code-based action (learned functions)} \\
\add{CoPiC~\cite{tian2026copic}} & \add{ALFWorld, NetHack, StarCraft II} & \add{Textual/symbolic input} & \add{Code-based action (program selection)} \\
\add{KnowAgent~\cite{zhu2025knowagent}} & \add{ALFWorld} & \add{Textual input} & \add{High-level action (knowledge-constrained)} \\
\add{MaestroMotif~\cite{klissarov2025maestromotif}} & \add{NetHack} & \add{Symbolic input} & \add{Code-based skill composition} \\
\add{ActiveVOO~\cite{liu2025activevoo}} & \add{ALFWorld} & \add{Visual input (active sensing)} & \add{High-level action} \\
\add{ReAd~\cite{zhang2024read}} & \add{Overcooked-AI} & \add{Symbolic input} & \add{High-level action (advantage feedback)} \\
\bottomrule
\end{tabular}}
\label{tab:agent_perception_action}
\vspace{-0.5cm}
\end{table}

\section{Multi-LLMGA Framework}
\label{sec:multi_agent_system}

In this section, we extend the single agent framework to multi-agent settings. Designing a multi-agent system in games is different from generic multi-agent systems because games impose unique constraints: game environments impose realistic constraints on information sharing: observations are partially observable and distributed across agents, and communication channels are often bandwidth-limited~\cite{zhang2023building}.

\begin{figure*}[htbp]
\centering
\includegraphics[width=14cm]{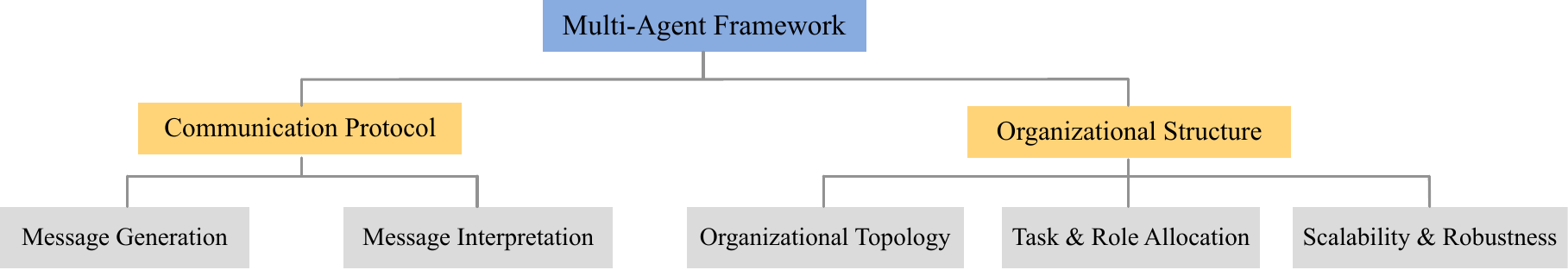}
\vspace{-0.2cm}
\caption{Overview of the multi-LLMGA framework.}
\vspace{-0.3cm}
\label{fig:mas_overview}
\end{figure*}

At the agent level, we examine how agents exchange information and integrate it into their decision-making. Communication protocols specify what messages are \textbf{generated} (e.g., observations, beliefs, or intentions) and how they are \textbf{interpreted} by receivers. At the organization level, we study three aspects: the \textbf{topology} of connections that shape communication flow, the \textbf{allocation of tasks and roles} that governs functional division of labor, and the mechanisms for ensuring \textbf{scalability and robustness} as groups expand. Figure~\ref{fig:mas_overview} presents the structure of this section of different components within the multi-LLMGA system.

\subsection{Communication Protocol}

In game and simulation environments, communication is likely constrained by partial observability, limited bandwidth, and asynchronous execution~\cite{zhang2023building}, which makes communication protocol design crucial for coordination. A communication protocol defines the rules that regulate peer-to-peer information exchange at the agent level, which specifies what message the sender should share, and how it is integrated by the receiver.

\textbf{Message Generation.}  Senders determine what type of information is worth exchanging, which can be broadly categorized into three classes: The first is \textit{observation}, referring to the raw and local signals each agent perceives from the environment. Observations are typically partial, such as perceiving only a limited visual field in the environment~\cite{zhang2023building}, sharing observations allows teammates to directly expand each other’s perceptual fields. Since raw perceptual input is often redundant or low-value, practical systems~\cite{zhang2023building} apply summarization to compress observations into compact, salient statements. The second is \textit{belief}, which represents an agent’s internal inference or probability distribution over the hidden state of the world, based on its own observations and prior knowledge~\cite{zhang2023proagent,multi_agent_coordination}. Compared to raw observations, beliefs provide higher-level interpretations. For example, an agent may observe scattered leaves and tree trunks, and infer that the environment contains sufficient wood resources nearby. The third is \textit{intention}, where agents communicate their planned actions or subgoals. Intention propagation is especially important in tasks that require complementary execution to reduce redundant effort (e.g., multiple agents pursuing the same subtask) and prevent conflicts (e.g., two agents competing for the same resource)~\cite{multi_agent_coordination,zhang2023building}.  In addition, when there is no communication mechanism/channel available, agents need to infer collaborators' hidden intentions based on their actions observed. \add{In adversarial games, senders also craft messages strategically rather than honestly: a study of Among Us finds impostor agents rely on equivocation rather than outright lies~\cite{milkowski2026amongus}, GPT-4o agents can out-deceive humans in Mafia~\cite{kao2025hidden}, and CoMet generates metaphor-based covert messages for the Undercover and Adversarial Taboo games~\cite{xu2025comet}. Persuasion and negotiation are likewise produced deliberately, as in Richelieu's goal-oriented negotiation grounded in social reasoning for Diplomacy~\cite{guan2024richelieu} and a Werewolf study of whether agents can generate opinion-leading speech~\cite{du2024helmsman}. To keep large-scale exchange efficient, EcoLANG induces a compact agent communication language~\cite{mou2025ecolang}.} 

\textbf{Message Interpretation.} Once communication takes place, agents need to integrate the exchanged information into their memory and ongoing decision process. In general, received messages can be directly adopted to guide actions. However, inconsistencies may arise when the new information conflicts with an agent’s existing internal state. To address this, agents must reconcile external messages and internal models. For instance, ProAgent~\cite{zhang2023proagent} infers the belief of a partner through the reasoning of theory of mind and subsequently corrects its estimate when the partner’s observed actions reveal mismatches. ReConcile~\cite{chen2024reconcile} provides a debate-based approach by engaging agents in multi-round discussions, where they attempt to convince each other with corrective explanations and aggregate responses through confidence-weighted voting to reach consensus. ECON~\cite{ECON} models this reconciliation as a Bayesian game, where agents treat each other’s beliefs and intentions as uncertain types and update them until they converge on a joint profile that all parties can consistently follow. \add{BEACOF likewise frames collaboration as a dynamic game of incomplete information, refining probabilistic beliefs about peers' capabilities until they reach an approximate perfect Bayesian equilibrium~\cite{fang2026belief}, and CSP4SDG interprets dialogue and game events in Avalon, Mafia, and Werewolf as evidence in a probabilistic constraint-satisfaction problem whose information-gain-weighted scores yield interpretable posterior beliefs over hidden roles~\cite{xu2025csp4sdg}.}

\subsection{Organizational Structure}

Organizational structure defines how agents are arranged and coordinated within a multi-agent system, including the topology of their connections, the allocation of tasks and roles, and the mechanisms that ensure scalability and stability as the population grows.  

\vspace{0.1cm}
\textbf{Organizational Topology.} Organizational topology refers to the structural constraints that determine how decisions flow, how agents connect for communication, and where authority over world state resides. Rather than a free design choice, topology is an architectural constraint that shapes the trade-off between scalability, robustness, and latency~\cite{qian2025scaling}.

\textit{Centralized} organization rely on a single planner or coordinator to aggregate information and allocate subtasks. This design ensures strong consistency and efficiency but creates bottlenecks and single points of failure, which limit scalability. For example, MindAgent~\cite{gong2023mindagent} adopts a single foundation model as the central dispatcher that issues the step-by-step commands to all agents. \textit{Decentralized} organization remove central authority and let agents act based on local observations and peer communication. Such topology is robust and can avoid global bottlenecks, but can suffer from coordination conflicts and redundant actions. CoELA~\cite{zhang2023building} follows this paradigm, framing cooperation as decentralized planning under costly communication channels. To balance coherence with local flexibility, \textit{hierarchical} organizations introduce multiple layers of control, with higher-level agents assigning goals or subtasks and lower-level agents refining them layer-by-layer. HAS~\cite{zhao2024hierarchical} exemplifies a three-tier hierarchy: a top-level manager sets global plans, intermediate conductors translate and distribute these plans, and bottom-level action agents execute concrete steps. Similarly, S-Agents~\cite{sagents} use a tree structure where a root node provides coordination and leaf nodes carry out subtasks. \textit{Partitioned} or sharded systems divide persistent environments into regions, each governed by local authority with cross-shard coordination handled by bridging protocols. This design enables scalability and fault tolerance, but weakens global consistency. Project Sid~\cite{sid} illustrates in a large-scale setting: thousands of Minecraft agents self-organize into civilizations where division of labor and institutions emerge, showing that centralized control is infeasible at such scale. \add{Game-focused frameworks add explicit structure on top of these topologies: HIMA organizes a society of specialized imitation-learning agents, each mastering a distinct StarCraft II tactic, under a strategic-planner meta-controller that synthesizes their proposals into coordinated actions~\cite{ahn2025hima}, and prompt-based organizational structures with designated leadership are imposed on embodied teams and iteratively refined through a criticize--reflect process to raise efficiency while lowering communication cost~\cite{guo2024organized}.}  

\vspace{0.1cm}
\textbf{Task \& Role Allocation.} Task and role allocation determines how subtasks are mapped to agents, shaping both efficiency and adaptability in multi-agent system. Allocation specifies the functional division of labor within the organizational topology. Three patterns are commonly observed: prefixed, dynamic, and emergent.

Prefixed allocation specifies roles or tasks in advance, often through a central planner or a leader. This ensures clear division of labor and prevents conflicts, making it reliable for structured environments but rigid under open-ended or rapidly changing conditions. MindAgent~\cite{gong2023mindagent} follows this approach: a single foundation model centrally dispatches per-step actions for all agents, directly specifying each agent’s next move. Similarly, S-Agents~\cite{sagents} predefine a root–leaf hierarchy, where the root serves as coordinator and leaves as executors, though the specific subtasks are still assigned dynamically during execution.  Dynamic allocation allows agents to determine their roles during execution, with assignments decided in real time by monitoring the environment or coordinating with peers. This increases adaptability and robustness but may produce redundancy when multiple agents converge on the same role. Overcooked-AI~\cite{carroll2019utility} illustrates this challenge, as frequent task changes require agents to split and reassign responsibilities on the fly. CoELA~\cite{zhang2023building} provides another example, where decentralized agents negotiate via natural language under costly communication channels to decide which subtasks to pursue. HAS~\cite{zhao2024hierarchical} also falls into this category: while roles such as manager and conductors are predefined, the system dynamically reorganizes action groups and reallocates responsibilities as tasks evolve.  \textit{Emergent} allocation does not predefine the set of roles but lets them arise through repeated interaction. 

At scale, Project Sid~\cite{sid} demonstrates how thousands of Minecraft agents spontaneously differentiate into specialized professions such as farmers, miners, builders, and traders, stabilizing cooperation without central control. This diversification arises from social awareness, where agents adjust goals in response to others’ activities, thereby reducing redundancy and enabling stable specialization. \add{For dynamic role coordination, COPPER assigns credit to individual agents via a shared reflector tuned with counterfactual rewards, producing role-personalized reflections that improve collaboration on tasks including chess~\cite{bo2024copper}.}

\textbf{Scalability \& Robustness.} Scaling multi-agent systems beyond small groups remains challenging. Early studies such as Generative Agents typically support only dozens of agents, since agents execute cognition through a sequential pipeline with a single thread. This serialized design becomes the bottleneck for scaling~\cite{park2023generative}. Project Sid addresses the per-agent bottleneck with the PIANO architecture, which runs six modules in parallel to update the agent state at different time scales. To prevent incoherence between simultaneous outputs, a cognitive controller~\cite{kaiya2023lyfe} selects an option from the candidate outputs of concurrent modules and transmits this decision to other modules for execution.

During the emergence of roles, certain factors are critical for ensuring organizational stability. Project Sid~\cite{sid} demonstrates that social awareness plays a critical role in sustaining division of labor: when agents observe many of their peers performing one task, they are more likely to select a different one. Through memory and repeated behavior, these roles become reinforced, allowing agents to form stable identities such as “farmer” or “miner” and yielding a more persistent specialization structure. In social simulation experiments, Artificial Leviathan~\cite{leviathan} demonstrate that memory depth is the key factor for the emergence of a commonwealth (i.e., the rise of a sovereign), under which social disorder is significantly reduced. This suggests that memory acts as a stabilizing mechanism by turning short-term interactions into long-term understanding of agents’ relative strengths and weaknesses, thereby forming group consensus. \add{Such coordination remains brittle at the level of collective behavior: an evaluation in the Melting Pot Commons Harvest game shows that LLM-augmented agents display a propensity for cooperation yet still fail to collaborate effectively under social dilemmas~\cite{mosquera2024meltingpot}.}

% \begin{table}[htbp]
% \centering
% \small
% \rowcolors{2}{gray!6}{white}
% \caption{Summary of representative LLMGAs in terms of Multi-LLMGA systems.}
% \resizebox{\textwidth}{!}{%
% \begin{tabular}{@{}ccll@{}}
% \toprule
% \textbf{Agent} & \textbf{Game / Environment} & \textbf{Communication Protocol} & \textbf{Organizational Structure} \\ 
% \toprule
% ProAgent~\cite{zhang2023proagent} &  & Theory-of-mind inference; belief correction & Decentralized coordination \\

% MindAgent~\cite{gong2023mindagent} &  & Centralized command dispatch & Centralized planner–executor \\

% CoELA~\cite{zhang2023building} &  & Natural language negotiation; costly channels & Decentralized planning \\

% % TeamCraft~\cite{long2024teamcraft} &  & Partial observability; coordination signals & Decentralized (prefixed roles) \\

% HAS~\cite{zhao2024hierarchical} &  & Hierarchical message passing & Hierarchical (manager–conductor–executor) \\

% S-Agents~\cite{sagents} &  & Goal propagation from coordinator to executors & Hierarchical tree structure \\

% Project Sid~\cite{sid} &  & Emergent communication; imitation and local sharing & Decentralized, partitioned society \\

% % Artificial Leviathan~\cite{leviathan} &  & Implicit communication via memory and actions & Decentralized emergent governance \\

% \bottomrule
% \end{tabular}}
% \label{tab:multi_llmga_summary}
% \end{table}

% \vspace{-0.2cm}
\section{Gameplay Taxonomy for LLMGA Design}
\label{sec:taxonomy}

The design of game agents is inseparable from the environments in which they operate: different genres foreground different capabilities, from fast perception–action cycles in action games to long-horizon planning in strategy games. A taxonomy that connects the properties of games with the demands they impose on agents is therefore valuable for this field. Here we adopt a challenge-centered game taxonomy: for each major category, we highlight the design challenge that most strongly drives LLMGA design. The genre axis itself draws on established categorizations, combining top-level groupings from SteamDB~\cite{steamdb2025} with the gameplay-oriented classification of Lee et al.~\cite{lee2014facet}. To make the taxonomy more coherent to covered studies, we exclude narrower genres such as driving/racing or fighting, and instead introduce sandbox as a category to capture open-ended and emergent play, with Minecraft as the canonical example.

Building on this taxonomy, we sketch how different game genres map into distinct design challenges. Action games require low-latency response, where agents are challenged to reconcile the slow deliberation of language models with the frame-level demands of real-time play. Adventure games highlight stateful world modeling, where progress depends on maintaining coherent memories of evolving environments, quests, and object dependencies. Role-playing games raise the issue of role fidelity, in which agents are expected to sustain consistent personas and align dialogue and actions with character identity. Strategy games emphasize opponent-aware planning, where the key difficulty lies in anticipating and adjusting to adversaries’ potential intentions under imperfect information. Simulation games emphasize dynamics fidelity, evaluating whether agents can exhibit behaviors that remain faithful to the governing simulation dynamics. Finally, sandbox games expose the challenge of open-ended goal progression, where agents are tasked with generating their own objectives, decomposing them hierarchically, and accumulating reusable skills to sustain long-term play.

\subsection{Action Games: Low-Latency Response}

\begin{wrapfigure}{r}{4.8cm}
\vspace{-1.0em}
\centering
\includegraphics[width=4.8cm]{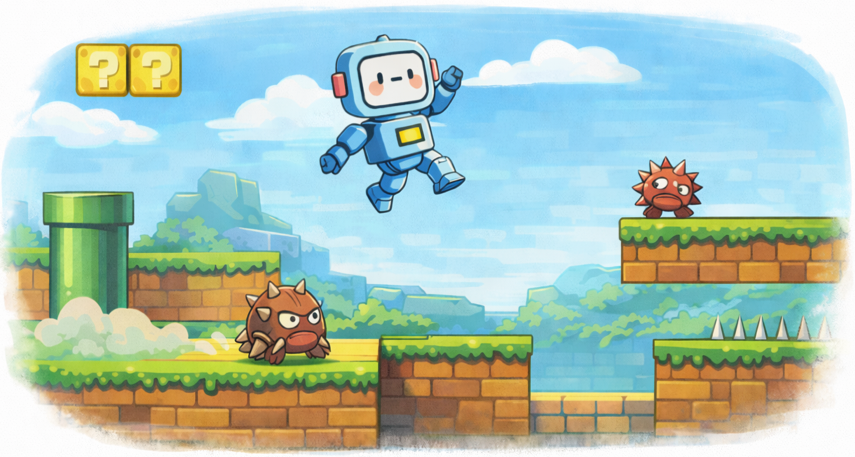}
\vspace{-0.7cm}
\caption{Action games require agents to respond with low latency and execute precise low-level control.}
\label{fig:action_game}
\vspace{-0.2cm}
\end{wrapfigure}

As shown in Figure~\ref{fig:action_game}, action games are characterized by real-time, time-critical interaction, where success hinges on executing precise movements such as aiming, dodging, or chaining combos within narrow temporal windows. This creates a fundamental demand for low-latency response, and the design challenge is therefore to reconcile the reasoning strengths of LLMs with the immediacy required by real-time gameplay.

\textbf{Environments.} 
Atari 2600 games in the Arcade Learning Environment~\cite{bellemare2013arcade} provide a canonical benchmark for reflexive control, where agents map raw pixel observations to joystick inputs at 60 Hz. Procgen~\cite{cobbe2020leveraging} extends this setup with procedurally generated levels, requiring agents to generalize their responses across unseen layouts rather than memorizing fixed patterns. 
Moving into 3D, ViZDoom~\cite{kempka2016vizdoom} present first-person 3D environments where perception is partial and high-dimensional, requiring agents to aim, strafe, and dodge in real time. 
Fighting games such as Street Fighter III~\cite{llm-colosseum} further sharpen the requirement for low-latency response: the timing of counters and combos is so precise that even minimal decision delays can flip the outcome of an exchange. 

\textbf{Methods.} Across action game environments, a consistent finding is that LLMs alone cannot sustain frame-level decision speed. Benchmarking across diverse video games including action games such as Super Mario and Street Fighter III shows that incorporating visual inputs often degrades rather than improves gameplay performance, partly because the additional processing overhead exacerbates inference latency~\cite{orak}. Further latency-sensitive evaluation on Street Fighter demonstrates that achieving competent play requires explicitly trading off reasoning quality for faster inference~\cite{kang2025win}.

To mitigate this bottleneck, researchers have adopted hybrid designs. One line of work delegates reflexive control to low-level policies trained through reinforcement or imitation learning, while reserving the LLM for high-level reasoning and strategy, as illustrated by two-tier agent systems in Overcooked~\cite{liu2023llm}. 
Empirical studies further show that latency-sensitive environments such as Street Fighter expose a fundamental trade-off between reasoning quality and decision speed: deeper reasoning produces stronger local decisions but increases inference latency to the point of losing more frequently, while shallower reasoning improves responsiveness and overall win rates~\cite{kang2025win}. \add{Action-game environments also serve as testbeds beyond latency: PoE-World learns a programmatic world model from few observations and embeds it into a model-based planner for stochastic Atari games such as Pong and Montezuma's Revenge~\cite{piriyakulkij2025poeworld}.}

% In the recent \textit{Black Myth: Wukong}, VARP samples frames at second-level intervals for multi-step action generation instead of conducting per-frame inference, thereby maintaining timely control under visually complex action settings~\cite{chen2024varp}.

% Research question: How to track the game state and understand the game world?
% Keyword: Game World Modeling, Memory System
% Keyword: State Tracking, Memory System

% \begin{wrapfigure}{r}{5cm}
% \vspace{-0.5em}
% \centering
% \includegraphics[width=5cm]{./Figure/adventure_game.pdf}
% \caption{xxx}
% \label{fig:adventure_game}
% \vspace{-0.5em}
% \end{wrapfigure}

\subsection{Adventure Games: Stateful World Modeling}

\begin{wrapfigure}{r}{4.5cm}
\vspace{-1.0em}
\centering
\includegraphics[width=4.5cm]{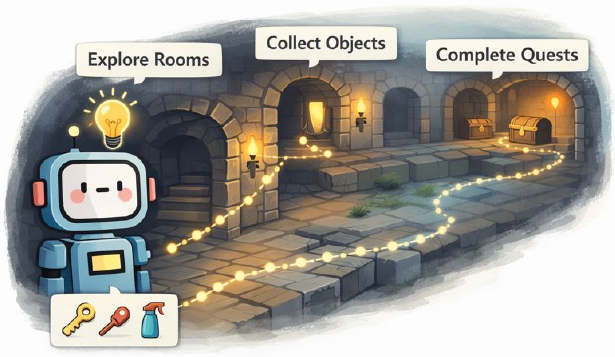}
\vspace{-0.7cm}
\caption{Adventure games typically involve exploration, object collection, and quest completion.}
\vspace{-0.2cm}
\label{fig:adventure_game}
% \vspace{-1.0em}
\end{wrapfigure}

As shown in Figure~\ref{fig:adventure_game}, adventure games are defined by partial observability and long-horizon quests: progress depends on remembering what has been explored, which preconditions of puzzles or storylines remain unsatisfied, and understanding how objects, actions, and the rules of the world interact. For LLMGAs, this creates a fundamental demand: they should be able to record, update, and retrieve both the evolving environment state and the underlying knowledge of how these elements can be used or combined. Without such modeling, agents lose track of progress, repeat past actions, or fail to connect prerequisites with goals. Empirically, GPT-3.5 struggles to construct coherent maps in partially known text-adventure environments, and state-prediction benchmarks indicate that even stronger LLMs are unreliable as implicit world simulators~\cite{hausknecht2020interactive}.

\textbf{Environments.} Adventure game benchmarks such as TextWorld~\cite{cote2019textworld}, Jericho~\cite{hausknecht2020interactive}, ALFWorld~\cite{shridhar2021alfworld}, and ScienceWorld~\cite{wang2022scienceworld} provide text-based environments in which players interact with the world through natural language, exploring rooms, collecting objects, and completing quests of varying complexity. 
For instance, TextWorld procedurally generates synthetic quests by varying the number of rooms, objects, and goals~\cite{TextWorldProblemsComp}. Jericho includes 56 human-authored classics such as the Zork series~\cite{zork1980,zorkIII1982} and Hitchhiker’s Guide to the Galaxy~\cite{HitchhikersTextAdventure30th}. ALFWorld aligns to the embodied ALFRED benchmark~\cite{shridhar2020alfred}, requiring agents to follow household instructions. ScienceWorld~\cite{wang2022scienceworld} simulates primary-school science curricula, highlighting basic knowledge from physics and chemistry for doing experiments.

\textbf{Methods.} Recent work has gradually converged on the view that memory should operate as the backbone of world modeling in adventure settings. Early agents such as ReAct~\cite{yao2022react} showed that simple interleaving of observations and actions is not sufficient, as the agent often fails to maintain an accurate view of the environment and becomes stuck. By incorporating reasoning, the agent can periodically summarize recent progress, ensuring that short-term records of explored locations, obtained items, and pending subgoals remain stable across steps. Reflexion~\cite{shinn2023reflexion} further demonstrates that writing self-critiques of failed attempts enables agents to extract insights from errors and avoid repeating them, thereby transforming episodic failures into persistent corrections of world knowledge. Subsequent agents, including Adapt~\cite{prasad2023adapt} and SwiftSage~\cite{lin2024swiftsage} further explicitly decompose quests into subgoals and track preconditions during execution. This keeps plans aligned with an evolving world state and enables coherent re-planning when branches fail. \add{In a similar spirit, LPLH equips interactive-fiction agents with structured map building and feedback-driven experience analysis to track and reuse world state more like human players~\cite{zhang2025lplh}.} KWM~\cite{qiao2025agentplanningworldknowledge} leverages successful trajectories to learn a knowledge-augmented world model, allowing agents to internalize regularities of environment dynamics and use the world model to guide future planning. \add{Beyond internalizing dynamics in memory, a recent line learns explicit, adaptive world models that predict future states: CoEx co-evolves a neurosymbolic world model alongside exploration so it does not go stale on ALFWorld and Jericho~\cite{kim2025coex}, DreamPhase plans by rolling out a latent world model offline with uncertainty-aware filtering~\cite{hamidi2026dreamphase}, and WorMI and TMoW retrieve and compose domain-specific world models at test time to adapt to new environments~\cite{yoo2025worldmodel,jang2026tmow}.} AriGraph~\cite{anokhin2024arigraph} encodes episodic experiences alongside semantic facts in a knowledge-graph memory, yielding a retrievable and interpretable representation of the game environment. \add{Such structured memory can also steer search: MC-DML couples Monte Carlo tree search with in-trial and cross-trial memory to dynamically reweight action evaluations in the Jericho benchmark~\cite{shi2025mcdml}.} At a larger scale, Cradle~\cite{tan2024towards} demonstrates the same principle in the visually rich adventure setting of Red Dead Redemption II, where the key difficulty lies in aligning perception with quest progress and narrative state. By maintaining memory as an explicit record of explored context and completed steps, Cradle enables the agent to keep exploration and story advancement coherent across long-horizon play, which stabilizes behavior in sprawling 3D environments. \add{Even with such mechanisms, world modeling stays hard in open-ended roguelikes: in NetHack, the zero-shot skill-based agent NetPlay tracks interaction history yet still struggles under sparse feedback and ambiguous goals~\cite{jeurissen2024nethack}.}

% \paragraph{Design takeaway.}
% Across these studies, the arc is clear: from short-term progress summaries, to long-term reflective updates, to subgoal and precondition tracking, to structured graphs and scalable multimodal modules—all converging on the same insight that \emph{memory is not auxiliary context but the agent’s world model}. Practically, an adventure agent should (i) keep a rolling snapshot in \emph{working memory} of local topology, interactable entities, and pending preconditions; (ii) \emph{write back} key events and dependencies into \emph{long-term memory} using structured stores (chunks, key–value, trees/graphs); and (iii) \emph{retrieve and check consistency} before acting, so that decisions remain faithful to the accumulated world state over extended play.

\subsection{Role-Playing Games: Role Fidelity}

\begin{wrapfigure}{r}{4.8cm}
\vspace{-1.0em}
\centering
\includegraphics[width=4.8cm]{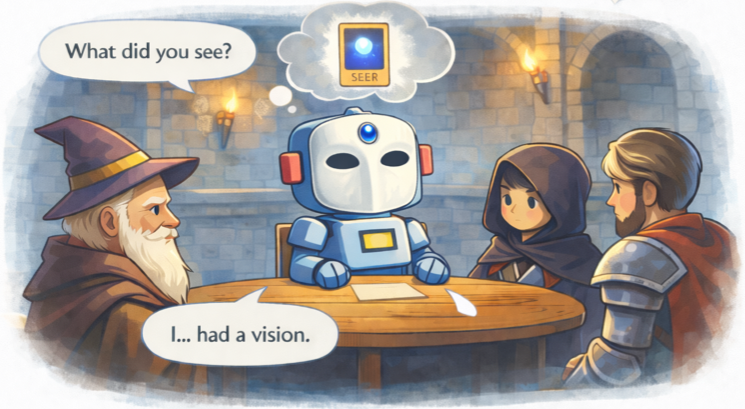}
\vspace{-0.7cm}
\caption{Role-playing games require agents to internalize and consistently enact pre-defined roles with distinct abilities, knowledge, and objectives.}
\vspace{-0.2cm}
\label{fig:rpg_game}
% \vspace{-1.0em}
\end{wrapfigure}

As illustrated in Figure~\ref{fig:rpg_game}, role-playing games (RPGs) require players to assume pre-defined characters with distinct abilities, knowledge, experiences, and objectives. Although RPGs may also incorporate elements of action or adventure, our focus here is on a common characteristic that underpins this genre: role fidelity. Role fidelity means that agents should internalize their assigned role and generate dialogue and actions that remain consistent with the character’s identity and capabilities. Failure to do so causes agents to lose consistency in speech and action, or even contradict their assigned role, undermining both immersion and gameplay effectiveness.

\textbf{Environments.} Social deduction board games provide natural testbeds for role fidelity. In Werewolf, each player receives a hidden role such as seer, guard, or werewolf, and must preserve secrecy while engaging in persuasion, deception, and coordinated voting~\cite{wereworlf1,werewolf_rl}. Similarly, Avalon assigns asymmetric roles with private knowledge (e.g., Merlin knowing the bad team), requiring agents to participate in multi-round discussions without revealing confidential information while still influencing team decisions~\cite{light2023avalonbench}. Negotiation games like Diplomacy, where each player embodies a nation with its own objectives~\cite{diplomacy}, and scripted murder-mystery games such as Jubensha~\cite{Wu2023Deciphering}, reinforce the same demand: agents must consistently inhabit a pre-defined persona and objectives, balancing what to disclose and what to withhold across multiple turns to preserve immersion and effectiveness.

Classic RPGs also emphasize role fidelity through long-horizon progression. For example, in Pokémon Red, the trainer role requires remembering the current storyline position, the Pokémon owned, the items carried, and the towns and paths visited. PokéAgent introduces exploration tasks to test whether agents can remain coherent as trainers throughout the gameplay~\cite{karten2025pokeagent}. Beyond the main character, role fidelity is even more critical for non-player characters (NPCs), which must sustain consistent personas across repeated interactions and emergent narratives, as exemplified by recent studies evaluating personality fidelity in role-playing~\cite{wang2024incharacter}.  

\textbf{Methods.} The simplest approach adds the role card directly into the prompt, listing traits and goals as initial memory~\cite{park2023generative}. While this establishes in-character openings, it quickly breaks down over multi-turn play, i.e., the role drift problem. Empirical studies show that in Avalon, LLMs may reveal their secret identity or fail to sustain deception across rounds~\cite{light2023avalonbench}. To mitigate such inconsistencies, approaches condition generation on explicit intentions or structured reasoning. For example, in Diplomacy, Cicero anchors dialogue in private strategic plans to ensure alignment between language and action~\cite{diplomacy}. These methods improve local consistency but are not designed to preserve long-term role fidelity. More recent approaches target role fidelity directly by integrating role profiles as a persistent component of the memory system. RoleLLM~\cite{wang2023rolellm} introduces structured role memory that separates private belief states (e.g., hidden identities) from public discourse records, ensuring that agents regulate what to disclose versus conceal across turns. CharacterLLM~\cite{shao2023character}, RoleLLM~\cite{wang2023rolellm} and CoSER~\cite{wang2025coser} adopt parametric adaptation, fine-tuning LLMs on curated role-play data to internalize persona traits and generate consistent style and objectives without continual reminders. These frameworks shift the focus from dialogue-level consistency to persistent memory management. \add{Role fidelity is also exercised in open-ended narrative and social play: the AI-native game 1001 Nights casts the player opposite an LLM-powered King in co-creative storytelling, where the model sustains narrative coherence and story keywords are materialized as in-game equipment via image generation~\cite{sun20231001nights}, while a multi-agent framework guides LLM agents through Avalon to study how they balance collaborative and confrontational behaviors and how this affects success rates~\cite{lan2024avalon}.}

\subsection{Sandbox Games: Open-Ended Goal Progression}

\begin{wrapfigure}{r}{4.5cm}
\vspace{-1.0em}
\centering
\includegraphics[width=4.6cm]{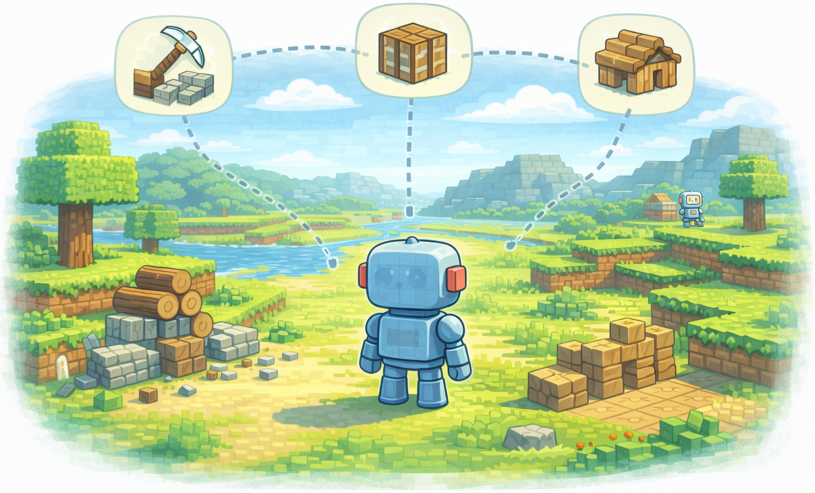}
\vspace{-0.7cm}
\caption{Sandbox games are typically open-ended, requiring agents to decide their own goals while freely exploring and building to support emergent progression.}
\vspace{-0.2cm}
\label{fig:sandbox_game}
% \vspace{-1.0em}
\end{wrapfigure}

Sandbox games are characterized by open-ended environments and emergent play rather than fixed quests or roles. As shown in Figure~\ref{fig:sandbox_game}, players can freely explore, collect resources, and set their own objectives from survival to large-scale construction. For LLMGAs, this creates unique demands for both generating meaningful goals in the absence of external instructions and decomposing goals into actionable plans. Without such mechanisms, agents either become stuck in aimless wandering or fail to coordinate long-horizon plans into coherent progression.

\textbf{Environments.} Minecraft and Crafter are two sandbox games that have been widely studied for game agents. Minecraft~\cite{minecraft} is a 3D sandbox game that offer players the great freedom to traverse a world made up of blocky, pixelated landscapes, facilitated by the procedurally generated worlds. The resource-based crafting system enables players to transform collected materials into tools, build elaborate structures and complex machines. Built on Minecraft, MineDojo~\cite{fan2022minedojo} provides a large-scale research platform with thousands of open-ended tasks, multimodal data from community sources, and the MineCLIP reward model. Crafter~\cite{hafner2021benchmarking} offers a lightweight 2D open-world environment with procedurally generated maps. It challenges players to manage their resources carefully to ensure sufficient water, food, and rest, while also defending against threats like zombies.

\textbf{Methods.} In sandbox settings, agents need to first determine what goals to pursue before they can decide how to achieve them. Existing works can be divided into two complementary directions. The first direction emphasizes goal generation through intrinsically motivated exploration. With LLMs, agents can propose adaptive goals conditioned on their current state, skills, and environment for curriculum learning. Voyager~\cite{wang2023voyager} exemplifies this idea by prompting an LLM to continually generate new objectives, building a self-directed curriculum and accumulating a library of reusable skills. OMNI~\cite{zhang2024omni} utilizes LLMs to determine interesting tasks for curriculum design, overcoming the previous challenge of quantifying "interestingness". ELLM~\cite{du2023guiding} queries LLMs for next goals given an agent’s current context, and rewards agents for accomplishing those suggestions in the sparse-reward setting; SPRING~\cite{wu2024spring} uses LLMs to summarize useful knowledge from the Crafter paper~\cite{hafner2021benchmarking} and progressively prompts the LLM to generate next action.

The second direction is hierarchical planning for task execution. Sandbox objectives such as constructing tools or building structures require agents to gather dispersed resources and follow multi-step recipes with strict dependencies. DEPS~\cite{wang2023describe} introduces plan correction: the LLM generates candidate subgoals, monitors execution outcomes, and self-explains failures in order to iteratively repair its plans, while leaving the final action execution to goal-conditioned controllers. Subsequent work emphasized making planning more reusable. JARVIS-1~\cite{wang2023jarvis} extend this idea by integrating multimodal perception and memory, grounding subgoal generation in visual context. Later work such as Plan4MC~\cite{yuan2023plan4mc} and RL-GPT~\cite{rlgpt} extend hierarchical planning by coupling high-level LLM planners with low-level controllers trained via reinforcement learning. Finally, multi-agent frameworks such as HAS~\cite{zhao2024hierarchical} and S-Agents~\cite{sagents} extend hierarchical planning to cooperative settings, dispatching subgoals across multiple agents to parallelize progress on complex objectives. \add{A third, increasingly direction grounds open-ended progression in learned world models. WALL-E aligns an LLM world model with the environment by extracting symbolic action rules and knowledge graphs into executable code, letting an RL-free model-predictive-control planner substantially raise success rates on the Minecraft-like Mars world~\cite{zhou2025walle}. ADAM autonomously learns a causal world graph from scratch, enabling interpretable lifelong task solving even in modified Minecraft worlds where prior crafting knowledge is unavailable~\cite{yu2025adam}. DLLM injects LLM-proposed subgoal hints into a model-based agent's imagined rollouts, rewarding hint-aligned transitions to improve exploration in the sparse-reward Crafter and Minecraft environments~\cite{liu2024dllm}.}

\vspace{-0.3cm}
\subsection{Strategy Games: Opponent-Aware Planning}

\begin{wrapfigure}{r}{4.5cm}
\vspace{-1.0em}
\centering
\includegraphics[width=4.5cm]{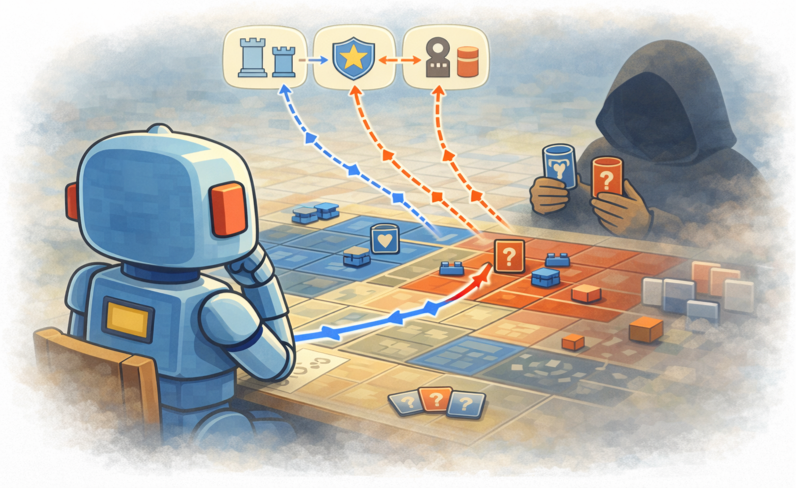}
\vspace{-0.7cm}
\caption{Strategy games emphasize strategic reasoning, requiring agents to anticipate opponent responses and plan over multiple future steps.}
\vspace{-0.2cm}
\label{fig:strategy_game}
% \vspace{-1.0em}
\end{wrapfigure}

Strategy games span a spectrum of complexity, from turn-based, deterministic, perfect information game to real-time, stochastic imperfect information games. As shown in Figure~\ref{fig:strategy_game}, a common requirement is opponent-aware planning: agents need to infer opponents’ possible intentions and conduct multi-step planning conditioned on these possibilities, as shown in Figure~\ref{fig:strategy_game}.

\textbf{Environments.} Board games like Chess and Go are fully observable, where agents need to search vast move trees while anticipating optimal counter-moves~\cite{feng2024chessgpt,toshniwal2022chess,OthelloGPT}. Pokémon battles~\cite{hu2024pokellmon} add uncertainty: players select moves or switches without knowing the opponent’s choice, and success depends on exploiting type matchups. 

Poker, exemplified by Texas Hold'em, deals each player two private hole cards,  followed by betting rounds as community cards are revealed. The winning strategy 
is not simply holding the best hand, but managing information asymmetry through bluffing, pot control, and reasoning about what cards the opponent may 
have~\cite{zhuang2025pokerbench}. StarCraft~II is a real-time strategy game where players collect resources, expand bases, build armies, and fight under the fog of war. Winning requires players to infer the opponent’s strategy from limited scouting, adapt build orders and timing attacks accordingly, and still control units precisely in battle. For agents, the challenge is therefore twofold: modeling and planning against an adaptive opponent as in other strategy games, and at the same time coordinating across macro, tactical, and micro levels under strict temporal constraints~\cite{starcraftii}.

\textbf{Methods.} In perfect-information games such as Chess and Go, opponent-aware planning reduces to deterministic search over long move sequences. ChessGPT~\cite{feng2024chessgpt} demonstrates that training on textual game corpora allows LLMs to evaluate positions and propose continuations, while blindfold-play studies~\cite{toshniwal2022chess,OthelloGPT} reveal that models can implicitly reconstruct board states from move sequences, approximating the effect of explicit lookahead search. \add{Building on this, MATE fine-tunes an LLM on a million chess positions annotated with expert strategy and tactic explanations, surpassing leading closed-source models at move selection~\cite{wang2025chessreasoning}. Explicit lookahead has also been layered on top of the model. Schultz et al. reach grandmaster-level chess by guiding Monte Carlo tree search externally or linearizing the game tree within the context window~\cite{schultz2025boardgames}. DiffuSearch instead replaces explicit search with discrete-diffusion implicit search over imagined future board states~\cite{ye2025diffusearch}.} In imperfect-information games, the challenge is reasoning over probability trees defined by partially observable states and uncertain opponent actions. Here, opponent modeling, often framed as theory-of-mind (ToM) thinking, is crucial. Suspicion-Agent~\cite{guo2023suspicion} shows that prompting LLMs for higher-order ToM in Leduc Hold’em leads to more aggressive raises and fewer passive calls, improving long-term chip gains. PokéLLMon~\cite{hu2024pokellmon,hu2025pokellmon} shows that LLM agents are still vulnerable to human misdirection strategies exploiting their limited higher-order ToM. For instance, a player may bait the agent by sending out a seemingly weak Pokémon, then switch to an immune one just before the attack lands, causing the agent to waste its move. \add{To strengthen opponent-aware play under such uncertainty, recent agents pair LLMs with lookahead and explicit opponent modeling. Pok\'eChamp embeds an LLM in minimax tree search with model-based opponent prediction for Pok\'emon battles~\cite{karten2025pokechamp}. Strategist couples high-level textual strategy generation with Monte Carlo tree search through self-play in the hidden-role games GOPS and Avalon~\cite{light2025strategist}. Such gains remain bounded by the models' inherent decision-making, as a systematic analysis across the dictator game, rock-paper-scissors, and a ring-network game finds persistent gaps from human rationality in forming preferences, refining beliefs, and acting optimally~\cite{fan2024rational}.}

% \textbf{Theory-of-Mind (ToM) Thinking.}ToM thinking is the ability to understand that others have mental states, such as beliefs and intentions~\cite{frith2005theory,kosinski2023theory}. Existing approaches prompt LLMs with specific questions to explicitly infer what others think (first-order) and what others think about its own thoughts (second-order). In imperfect information games like Leduc Hold’em, Suspicion-Agent~\cite{guo2023suspicion} demonstrates that higher-order ToM lead agents to adopt more aggressive strategies (higher Raise rate) and reduce passive following (lower Call rate), and thus can accumulate more chips. In collaboration games such as Overcooked, ToM thinking enables LLM agents to adjust their actions to adapt to their partners' actions and offer assistance when necessary~\cite{multi_agent_coordination}. ProAgent~\cite{zhang2023proagent} introduces a belief revision module to revise the wrong belief toward their partners when they observe partners' new actions; Li et al.~\cite{li2023theory} demonstrates that maintaining a belief state toward teammates' past observations and task-relevant objects can significantly improve ToM capabilities by reducing hallucination.

\subsection{Simulation Games: Dynamics Fidelity}

\begin{wrapfigure}{r}{4.5cm}
\vspace{-1.0em}
\centering
\includegraphics[width=4.5cm]{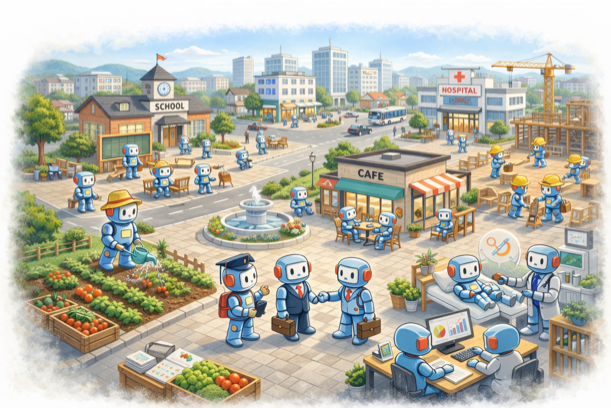}
\vspace{-0.7cm}
\caption{Simulation games aim to model or reproduce the behavior of a system and require agents to behave consistently with the underlying dynamics of the simulation.}
\label{fig:simulation_game}
\vspace{-1.0em}
\end{wrapfigure}

As illustrated in Figure~\ref{fig:simulation_game}, simulation games model the behavior of a system governed by dynamics rules such as social norms, economic mechanisms, ecological constraints, or physical laws. Rather than focusing on fixed objectives, these environments emphasize the faithful operation of the underlying simulation processes, \textit{i.e.}, dynamics. We therefore characterize the core challenge as dynamics fidelity, referring to an agent’s ability to generate behaviors that drive the state transitions of the simulated system in a manner consistent with its governing dynamics.

\textbf{Environments.}
Human simulation environments construct virtual societies for studying emergent social behavior. 
Generative Agents~\cite{park2023generative} places 25 agents in a sandbox town with cognitive modules for everyday interaction. 
Humanoid Agents~\cite{wang2023humanoid} extends this setting by incorporating physiological needs, emotions, and relationship closeness. Beyond human simulation, CivRealm~\cite{qi2024civrealm} is a Civilization-style simulation environment focusing on the macro-scale evolution of societies across historical eras. \add{From a generative angle, Unbounded realizes an infinite character life-simulation game whose mechanics, narrative, and interactions are produced on the fly by a distilled LLM with consistent visual character generation~\cite{li2024unbounded}.} 

% More recent platforms such as Project Sid~\cite{sid} scale to hundreds or thousands of agents in a Minecraft-based world, while Artificial Leviathan~\cite{leviathan} creates a survival sandbox for exploring the emergence of social contracts and authority. 

\textbf{Methods.} For social simulation games, maintaining dynamics fidelity in simulation requires that agents behave in ways consistent with human or societal patterns rather than drifting into unrealistic behavior. 
Generative Agents~\cite{park2023generative} achieved this by introducing cognitive architectures with memory, reflection, and planning. Its memory system scores experiences by recency, relevance, and importance, allowing salient events to be repeatedly recalled and consolidated, mirroring core patterns of human memory. Humanoid Agents~\cite{wang2023humanoid} further improved fidelity by embedding physiological needs, emotions, and relationship closeness into decision-making, leading agents to display more human-like variability. \add{Similarly, a desire-driven agent selects activities to satisfy multi-dimensional intrinsic needs, producing coherent human-like daily routines in the Concordia simulator~\cite{wang2025desiredriven}.} At larger scales, Project Sid~\cite{sid} constructed an agent society in a Minecraft-based world, inhabited by hundreds to thousands of agents who shared limited resources and interacted concurrently. Under conditions of scarcity and continual co-presence, the agents competed and cooperated, spontaneously developing specialized roles, adapting collective rules, and propagating cultural practices such as religion. Artificial Leviathan~\cite{leviathan} approaches fidelity through a survival sandbox in which agents, driven by psychological needs under resource pressure, choose among farming, trading, or robbing each day. Empirical results show that agents start in conflict but eventually form social contracts, authorize a sovereign, and transition to peaceful cooperation. Experiments further show that parameters like memory depth has a large impact on the speed and nature of social evolution. \add{Such emergent dynamics also surface in game economies, where a generative agent-based model reproduces role specialization and realistic market dynamics in massively multiplayer games~\cite{xu2025mmoeconomy}.} \add{However, a distinct concern is whether it is genuinely faithful to real humans rather than merely plausible. Validation against established human findings is encouraging: given heterogeneous personas and theory of mind, LLM agents reproduce human third-party punishment in public-goods games~\cite{cross2025normgabm}, match human choices in the behavioral-economics Trust Game~\cite{xie2024trust}, and recover the predictions of Homans' social exchange theory~\cite{wang2025homans}. To close residual gaps, other work calibrates simulations to empirical data, aligning population-level decision dynamics to real social statistics~\cite{mi2025mfllm} or tuning agent personas until crowd behavior matches expert benchmarks~\cite{wang2025peba}. Diagnostic studies nonetheless caution that human-like socialization may not genuinely emerge in such open-ended agent societies~\cite{li2026moltbook}.}

\section{Discussion and Open Challenges}
\label{sec:discussion}

This section synthesizes the discussions from previous sections and connects them to the two research questions: (Q1) how the core architectural components of LLMGAs are designed, and (Q2) how genre-specific requirements shape these architectural components. We also highlight open challenges and future research directions.

\subsection{Memory System}

To answer Q1, we introduced memory systems as a core architectural component for storing, retrieving, and recalling past experiences in order to maintain behavioral consistency and inform decision-making. The memory system is divided into working memory and long-term memory, serving distinct yet interdependent functions (§\ref{sec:memory}). Working memory stabilizes short-horizon decision-making under limited context by (i) extending the effective input span, (ii) compressing redundant information, and (iii) maintaining recent bindings and plans. These mechanisms mitigate short-term drift and prevent inconsistent actions~\cite{hu2025pokellmon}. Long-term memory ensures continuity across episodes when organized into structured and retrievable forms such as chunks with metadata, key–value pairs, hierarchical trees, graphs, or skill libraries. The interaction between the two hinges on three functions: consolidation, which determines when transient traces are committed to durable storage; structuring, which organizes stored content for efficient access; and retrieval, which reactivates relevant information through metadata filtering, semantic search, or traversal of structured memories. 

An open challenge for current memory systems is to move beyond “storing more” toward developing a true “world-model” memory that consolidates fragmented experiences into a coherent mental model of the game world~\cite{johnson1983mental}. This limitation is particularly pronounced in adventure games, where long-horizon exploration and partial observability place strong demands on persistent world modeling (Q2). To obtain a world-model memory system and distinguish it from a mere database, three design principles can be essential: (i) Predictive dynamics: memory should not only replay past events but also predict what might happen next. In cognitive science, mental models are understood as internal simulations that help people anticipate outcomes and detect errors, rather than as static records~\cite{johnson1983mental}. (ii) Structural compositionality: experiences need to be stored in organized forms, such as schemas or graphs that link entities, relations, and precondition–effect rules, so that knowledge from different situations can be combined and reused. This idea aligns with schema and situation-model theories, which show that humans build integrated “who–what–where–when–why” representations to reason beyond literal experiences~\cite{zwaan1998situation}. (iii) Selective consolidation and forgetting: long-term memory should decide what to keep and what to discard. Instead of saving every detail, it should preserve experiences that are important for understanding or improving the current model of the world, while letting irrelevant or low-value details fade. Research on human memory shows that people tend to remember information that is useful or frequently encountered and forget what rarely matters~\cite{anderson1991reflections}. 

% Together, these principles describe how a world model maintains coherence and usefulness over time by keeping and updating memories that improve prediction, organizing them into meaningful structures, and forgeting what no longer important.

\subsection{Reasoning Mechanism}

Reasoning in LLMGAs is not merely about generating intermediate thoughts, but also ensuring that these thoughts lead to correct decisions (§\ref{sec:reasoning}). To answer Q1, we categorize existing approaches into instruction-guided reasoning and fine-tuning–based reasoning enhancement. Instruction-guided reasoning highlights recurring challenges: reasoning should avoid error propagation and remain consistent across steps. In comparison, fine-tuning approaches improve reasoning abilities by grounding reasoning in experience and feedback. Despite these advances, a fundamental limitation remains: current approaches rely on narrow forms of feedback or numeric rewards. Multi-path reasoning improves robustness by exploring diverse reasoning trajectories, yet it provides no learning signal about which paths are preferable or why. Reflective reasoning enables self-correction across episodes but remains coarse-grained, offering post-hoc summaries rather than actionable, step-level feedback. Process Reward Models (PRMs) attempt to provide this supervision by assigning stepwise rewards, but rely heavily on costly human annotation or handcrafted heuristics, making feedback sparse, rigid, and poorly aligned with the linguistic nature of reasoning.

The deeper challenge lies in the mismatch between the form of reinforcement and the medium of reasoning. Traditional reinforcement learning depends on numeric rewards, whereas reasoning in LLMs unfolds through language, where success, failure, and state changes appear as semantic cues. Humans, however, are capable of assigning credit even from weak or indirect feedback: they adjust their reasoning based on partial signals such as environmental changes, the outcome of intermediate goals, or the perceived coherence of an explanation. Cognitive studies on metacognition and error monitoring show that such internal evaluation enables people to refine reasoning continuously through semantic and contextual signals rather than explicit numeric reinforcement~\cite{yeung2012metacognition,beran2012foundations,fleming2012neural}. By virtue of their linguistic grounding, LLMs can transform textual feedback, environmental descriptions, and self-critiques into implicit reinforcement signals, generalizing traditional reward learning beyond numeric values and enabling reasoning to improve through understanding rather than scoring. As for Q2, across strategy games, opponent-aware planning poses a central reasoning challenge. In deterministic, perfect-information games, the main difficulty lies in evaluating long sequences of alternating moves to anticipate the opponent’s best actions. In imperfect-information games, agents must reason under uncertainty by forming and updating beliefs about hidden states and possible opponent actions, which often requires higher-order theory-of-mind reasoning to explicitly model the opponent’s intentions, predict their likely responses, and adjust strategies accordingly over multiple steps.

\subsection{Perception-Action Interface}

% From the perspective of RQ1, the perception \& action interface grounds how agents see the environment and fulfill their decisions (§\ref{sec:in/output}). {\color{deepblue}
% Across the surveyed literature, perception interfaces tend to be most effective when they emphasize decision-relevant features, such as object states, affordances, and strategic cues. Conversely, existing action interfaces reflect a recurring trade-off between expressivity and reliability: high-level actions simplify the decision space, low-level controls enable fine-grained execution, and programmatic actions provide structure, verifiability, and reusability.
% }

% A key challenge is how effectively perception and action are aligned to support decision quality. Perception should highlight decision-relevant features such as object states, affordances, and strategic cues so that the agent does not waste capacity on irrelevant detail. Action interfaces, in turn, balance expressivity and reliability: high-level actions simplify decision space, low-level controls allow fine precision, and programmatic actions offer structure, verifiability, and reusability. Overall, perception and action should be co-designed as a coupled system, since they form a single loop where perception shapes possible actions and actions in turn shape what must be perceived. Ensuring this alignment while keeping the loop efficient and scalable remains an open problem for future research.

% ----------

From the perspective of Q1, the perception \& action interface grounds how agents see the environment and fulfill their decisions (§\ref{sec:in/output}). 
Across the surveyed literature, perception interfaces tend to be most effective when they emphasize decision-relevant features, such as object states, affordances, and strategic cues, rather than exhaustively encoding raw observations.
However, increasing perceptual richness often comes at the cost of efficiency, as more complex representations incur higher preprocessing and inference overhead. A similar tension arises on the action side. Existing action interfaces reflect a recurring trade-off between effectiveness and efficiency: high-level actions simplify the decision space and stabilize reasoning but may limit control precision, while low-level controls enable fine-grained execution but require high-frequency generation.

For Q2, the challenge of low-latency response is particularly pronounced in action games, where agents operate under strict real-time constraints while issuing low-level control signals at high frequency. In such environments, even strong strategic reasoning can fail if actions are generated too slowly.
These constraints expose a requirement for hierarchical action generation. Low-level action execution is typically lightweight and reactive, while higher-level strategic reasoning operates at a slower timescale.
A practical direction for future research is therefore to explicitly decouple strategic decision-making from reflexive control, allowing LLMs to generate abstract intents or subgoals, which are then executed by fast, specialized controllers.
Further opportunities include adaptive abstraction, where the granularity of perception and action is dynamically adjusted based on latency budgets and task demands.
Designing perception \& action interfaces that jointly optimize effectiveness and efficiency remains an open challenge for LLMGA research.

\subsection{Multi-LLMGA System}

LLM-based multi-agent systems extend game environments from single-agent decision making to collective behavior, introducing new challenges such as communication bandwidth limits, and the need to preserve realistic interaction constraints (§\ref{sec:multi_agent_system}).  To answer Q1, we analyzed these systems across two complementary levels. At the agent level, communication protocols determine what information agents exchange and how it is integrated under these constraints, while at the organization level, organizational structures govern decision flow (topology), guide division of labor (task allocation), and determine whether societies can scale and remain stable.

For Q2, we are interested in how genre-specific requirements influence the multi-agent architectural design. Role-playing games such as Werewolf and Avalon center on agent-level communication: the core challenge is controlling what to disclose and what to withhold, placing the design emphasis on belief sharing and selective information exchange. Strategy games such as StarCraft~II and Overcooked also foreground agent-level coordination, where the focus shifts to rapid intention propagation and conflict avoidance. In contrast, simulation and sandbox games operating at larger scales, such as those modeling entire virtual societies in Minecraft, shift the architectural focus from agent-level protocols to organization-level structures. As the number of agents grows, the primary challenge becomes designing topologies and role allocation mechanisms that sustain stable coordination without centralized control.

Prior studies have demonstrated the potential of multi-agent systems in large-scale simulations, where agents exhibit emergent behaviors. However, current large-scale multi-agent simulations remain constrained by structural and methodological limitations. Many “emergent” phenomena, such as role differentiation, norm formation, or collective planning, are closely tied to task initialization and rule design. In practice, agents are often seeded with shared goals, cooperation-oriented prompts, or predefined role templates that guide subsequent division of labor and coordination patterns. Prior studies of multi-agent societies have shown that such structural priors are widespread, from small-scale social environments~\cite{park2023generative} to hierarchical and large-scale simulations~\cite{sagents,zhao2024hierarchical}, where coordination often reflects the constraints of task setup rather than fully autonomous self-organization. Moreover, the lack of open and reproducible large-scale platforms further limits systematic evaluation, making it difficult to test under what specific conditions such collective dynamics genuinely arise.

\subsection{Game Environments and Benchmarks}

Table~\ref{tab:benchmarks} summarizes existing open-sourced benchmarks or environments for LLMGAs. Some widely used benchmarks (\textit{e.g.}, TextWorld~\cite{cote2019textworld}, ALFWorld~\cite{shridhar2021alfworld}, ScienceWorld~\cite{wang2022scienceworld}) were primarily developed before the rise of LLMs. Their tasks are generated from templated rules and constrained by a limited set of admissible actions and shallow dynamics, which result in highly similar instantiated tasks and low interactive complexity. In ALFWorld, for example, tasks are constructed from household instruction templates over a fixed action set (e.g., pick up, open, put, heat), producing many near-duplicate instances that only substitute objects or receptacles~\cite{shridhar2021alfworld}. 

% Likewise, ScienceWorld defines 30 scientific task types instantiated into 1,483 training and 241 test tasks within a single lab-style domain, with substantial overlap in objects and interaction rules~\cite{wang2022scienceworld}. 

\begin{table}[htbp]
\centering
% \footnotesize
\scriptsize
\caption{Open-sourced Benchmark/Environments for LLMGAs}
\vspace{-0.2cm}
\resizebox{\textwidth}{!}{%
\begin{tabular}{@{}lllllll@{}}
\toprule
\textbf{Genre} & \textbf{Benchmark/Environment} & \textbf{Game Content} & \textbf{Player Mode} & \textbf{Modality} & \textbf{Code Link} & \textbf{Date} \\ \midrule

% ===================== Action Games =====================
\multirow{4}{*}{Action}
& Overcooked-AI~\cite{carroll2019utility} & Overcooked-like Game & Multi & Symbolic & \href{https://github.com/HumanCompatibleAI/overcooked_ai}{GitHub} & 2019/12 \\
& LLM-Coordination~\cite{multi_agent_coordination} & Overcooked-AI & Multi & Text & \href{https://github.com/eric-ai-lab/llm_coordination}{GitHub} & 2023/10 \\
& TextStarCraft~\cite{starcraftii} & StarCraft II & Single & Text & \href{https://github.com/histmeisah/Large-Language-Models-play-StarCraftII}{GitHub} & 2023/12 \\
& llm-colosseum~\cite{llm-colosseum} & Street Fighter III & Multi & Vision & \href{https://github.com/OpenGenerativeAI/llm-colosseum}{GitHub} & 2024/03 \\
\midrule

% ===================== Adventure Games =====================
\multirow{7}{*}{Adventure}
& VirtualHome~\cite{puig2018virtualhome} & Household Tasks & Single & Mixed & \href{https://github.com/xavierpuigf/virtualhome}{GitHub} & 2018/06 \\
& TextWorld~\cite{cote2019textworld} & Text-based Games & Single & Text & \href{https://github.com/microsoft/TextWorld}{GitHub} & 2018/07 \\
& Jericho~\cite{hausknecht2020interactive} & Interactive Fictions & Single & Text & \href{https://github.com/microsoft/jericho}{GitHub} & 2019/09 \\
& ALFRED~\cite{shridhar2020alfred} & Household Tasks & Single & Mixed & \href{https://github.com/askforalfred/alfred}{GitHub} & 2020/03 \\
& ALFWorld~\cite{shridhar2021alfworld} & Household Tasks & Single & Text & \href{https://github.com/alfworld/alfworld}{GitHub} & 2020/10 \\
& ScienceWorld~\cite{wang2022scienceworld} & Science Experiments & Single & Text & \href{https://github.com/allenai/ScienceWorld}{GitHub} & 2022/03 \\
& BabyAI-Text~\cite{carta2023grounding} & MiniGrid Tasks & Single & Text & \href{https://github.com/mila-iqia/babyai}{GitHub} & 2023/02 \\
\midrule

% ===================== Role-playing Games =====================
\multirow{11}{*}{Role-playing}
& Cicero~\cite{diplomacy} & Diplomacy & Multi & Text & \href{https://github.com/facebookresearch/diplomacy_cicero}{Github} & 2022/12 \\
& Generative Agents~\cite{park2023generative} & Sims-like Game & Multi & Text & \href{https://github.com/joonspk-research/generative_agents}{GitHub} & 2023/04 \\
& AgentSims~\cite{lin2023agentsims} & Sims-like Game & Multi & Text & \href{https://github.com/py499372727/AgentSims}{GitHub} & 2023/08 \\
& Xu et al.~\cite{wereworlf1} & Werewolf & Multi & Text & \href{https://github.com/xuyuzhuang11/Werewolf}{GitHub} & 2023/09 \\
& AvalonBench~\cite{light2023avalonbench} & Avalon & Multi & Text & \href{https://github.com/jonathanmli/Avalon-LLM}{GitHub} & 2023/10 \\
& Humanoid Agents~\cite{wang2023humanoid} & Sims-like Game & Multi & Mixed & \href{https://github.com/HumanoidAgents/HumanoidAgents}{GitHub} & 2023/10 \\
& \add{clembench~\cite{chalamalasetti2023clembench}} & \add{Dialogue Games} & \add{Multi} & \add{Text} & \add{\href{https://github.com/clp-research/clembench}{GitHub}} & \add{2023/11} \\
& \add{SOTOPIA~\cite{zhou2023sotopia}} & \add{Social-interaction Scenarios} & \add{Multi} & \add{Text} & \add{\href{https://github.com/sotopia-lab/sotopia}{GitHub}} & \add{2023/10} \\
& PokéAgent Challenge~\cite{karten2025pokeagent} & Pokémon & Single & Text & \href{https://github.com/sethkarten/pokeagent-speedrun}{GitHub} & 2025/07 \\
& \add{RoleLLM~\cite{wang2023rolellm}} & \add{Role-Playing Benchmark} & \add{Single} & \add{Text} & \add{\href{https://github.com/InteractiveNLP-Team/RoleLLM-public}{GitHub}} & \add{2023/10} \\
& \add{CoSER~\cite{wang2025coser}} & \add{Role-Playing Benchmark} & \add{Multi} & \add{Text} & \add{\href{https://github.com/Neph0s/CoSER}{GitHub}} & \add{2025/02} \\
\midrule

% ===================== Strategy Games =====================
\multirow{11}{*}{Strategy}
& Cicero~\cite{diplomacy} & Diplomacy & Multi & Text & \href{https://github.com/facebookresearch/diplomacy_cicero}{Github} & 2022/12 \\
& Xu et al.~\cite{wereworlf1} & Werewolf & Multi & Text & \href{https://github.com/xuyuzhuang11/Werewolf}{GitHub} & 2023/09 \\
& AvalonBench~\cite{light2023avalonbench} & Avalon & Multi & Text & \href{https://github.com/jonathanmli/Avalon-LLM}{GitHub} & 2023/10 \\
& TextStarCraft~\cite{starcraftii} & StarCraft II & Single & Text & \href{https://github.com/histmeisah/Large-Language-Models-play-StarCraftII}{GitHub} & 2023/12 \\
& PokéLLMon~\cite{hu2025pokellmon} & Pokémon Battles & Single & Text & \href{https://github.com/git-disl/PokeLLMon}{GitHub} & 2024/02 \\
& PokerBench~\cite{zhuang2025pokerbench} & Texas Hold'em & Single & Text & \href{https://github.com/pokerllm/pokerbench}{GitHub} & 2025/01 \\
& ChessGPT~\cite{feng2024chessgpt} & Chess & Single & Symbolic & \href{https://github.com/waterhorse1/ChessGPT}{GitHub} & 2024/03 \\
& CivRealm~\cite{qi2024civrealm} & Civilization-like Game & Single & Symbolic & \href{https://github.com/bigai-ai/civrealm}{GitHub} & 2024/01 \\
& \add{LMAct~\cite{ruoss2025lmact}} & \add{Chess, Atari, Tic-tac-toe} & \add{Single} & \add{Mixed} & \add{\href{https://github.com/google-deepmind/lm_act}{GitHub}} & \add{2025/01} \\
& \add{LLM-PySC2~\cite{li2025pysc2}} & \add{StarCraft II} & \add{Multi} & \add{Mixed} & \add{\href{https://github.com/NKAI-Decision-Team/LLM-PySC2}{GitHub}} & \add{2025/04} \\
& \add{Full-press Diplomacy~\cite{duffy2025diplomacy}} & \add{Diplomacy} & \add{Multi} & \add{Text} & \add{\href{https://github.com/GoodStartLabs/AI_Diplomacy}{GitHub}} & \add{2025/08} \\
\midrule

% ===================== Simulation Games =====================
\multirow{5}{*}{Simulation}
& MineDojo~\cite{fan2022minedojo} & Minecraft & Single & Mixed & \href{https://github.com/MineDojo/MineDojo}{GitHub} & 2022/06 \\
& Generative Agents~\cite{park2023generative} & Sims-like Game & Multi & Text & \href{https://github.com/joonspk-research/generative_agents}{GitHub} & 2023/04 \\
& AgentSims~\cite{lin2023agentsims} & Sims-like Game & Multi & Text & \href{https://github.com/py499372727/AgentSims}{GitHub} & 2023/08 \\
& Humanoid Agents~\cite{wang2023humanoid} & Sims-like Game & Multi & Mixed & \href{https://github.com/HumanoidAgents/HumanoidAgents}{GitHub} & 2023/10 \\
& CivRealm~\cite{qi2024civrealm} & Civilization-like Game & Single & Symbolic & \href{https://github.com/bigai-ai/civrealm}{GitHub} & 2024/01 \\
\midrule

% ===================== Sandbox Games =====================
\multirow{7}{*}{Sandbox}
& Crafter~\cite{hafner2021benchmarking} & 2D Survival Sandbox & Single & Vision & \href{https://github.com/danijar/crafter}{GitHub} & 2021/06 \\
& MineDojo~\cite{fan2022minedojo} & Minecraft & Single & Mixed & \href{https://github.com/MineDojo/MineDojo}{GitHub} & 2022/06 \\
& Odyssey~\cite{liu2024odyssey} & Minecraft & Single & Mixed & \href{https://github.com/zju-vipa/Odyssey}{GitHub} & 2024/07 \\
& Mars~\cite{tang2024mars} & Crafter & Single & Vision & \href{https://github.com/XiaojuanTang/Mars}{GitHub} & 2024/10 \\
& \add{Baba Is AI~\cite{cloos2024babaisai}} & \add{Rule-manipulation Puzzle} & \add{Single} & \add{Vision} & \add{\href{https://github.com/nacloos/baba-is-ai}{GitHub}} & \add{2024/07} \\
& \add{Plancraft~\cite{dagan2025plancraft}} & \add{Minecraft} & \add{Single} & \add{Mixed} & \add{\href{https://github.com/gautierdag/plancraft}{GitHub}} & \add{2025/03} \\
& \add{UnrealZoo~\cite{zhong2025unrealzoo}} & \add{Photo-realistic 3D Worlds} & \add{Multi} & \add{Vision} & \add{\href{https://github.com/UnrealZoo/unrealzoo-gym}{GitHub}} & \add{2025/06} \\
% & TeamCraft~\cite{long2024teamcraft} & Minecraft & Multi & Mixed & \href{https://github.com/teamcraft-bench/TeamCraft}{GitHub} & 2024/12 \\
\midrule

% ===================== Diverse =====================
\multirow{5}{*}{Diverse}
& CuisineWorld~\cite{gong2023mindagent} & Cooperative Tasks & Multi & Text & \href{https://github.com/mindagent/mindagent}{GitHub} & 2023/09 \\
& Cradle~\cite{tan2024towards} & Multiple Video Games & Single & Mixed & \href{https://github.com/BAAI-Agents/Cradle}{GitHub} & 2024/03 \\
& BALROG~\cite{paglieri2025balrog} & Multiple RL Games & Single & Mixed & \href{https://github.com/balrog-ai/BALROG}{GitHub} & 2024/11 \\
& lmgame-Bench~\cite{lmgame-bench} & Multiple Video Games & Single / Multi & Mixed & \href{https://github.com/lmgame-org/GamingAgent}{GitHub} & 2025/05 \\
& Orak~\cite{orak} & Multiple Video Games & Single & Mixed & \href{https://github.com/krafton-ai/Orak}{GitHub} & 2025/06 \\
\bottomrule
\end{tabular}}
\vspace{-0.3cm}
\label{tab:benchmarks}
\end{table}

High-quality game environments/benchmarks are crucial for advancing the capabilities of LLMGAs. Such environments should not only be more complex, but complex in targeted ways that expose the distinctive weaknesses of current architectures. This entails: (i) tasks with deeper compositional structure and long-horizon dependencies, ensuring that success cannot be reduced to pattern-matching templates; (ii) world dynamics governed by consistent physical or social rules, requiring agents to acquire and exploit regularities rather than memorize isolated instances; and (iii) scalability in both breadth (diverse tasks and domains) and depth (persistent settings spanning multiple days or large populations of agents).

Most existing benchmarks evaluate game agents with coarse-grained metrics such as win rate and task success rate~\cite{shridhar2021alfworld,wang2022scienceworld}. While these high-level measures capture overall gameplay performance, they obscure where and why agents fail. Moving forward, the field requires fine-grained evaluation protocols that can diagnose the core components of agent design, memory, reasoning, perception–action translation, and multi-agent coordination, thus linking empirical evaluation to theoretical progress. One practical approach is game-specific metric design. Such metrics leverage domain knowledge to expose failure modes that aggregate scores cannot reveal. For example, PokéLLMon introduces the consecutive switch rate, measuring the proportion of turns where the agent switches Pokémon consecutively as a proxy for short-term inconsistency~\cite{hu2024pokellmon}. Voyager uses map coverage and number of unique items collected to quantify exploration breadth and inventory management~\cite{wang2023voyager}. At a larger scale, Project Sid~\cite{sid} invite new metrics, such as persistence of social norms or stability of emergent institutions, providing outcome measures with diagnostic signals for interpreting agent behavior.

However, not all evaluation targets lend themselves to direct quantification. Aspects such as role fidelity, believability, or the coherence of emergent behavior often require judgment-based protocols. In Generative Agents~\cite{park2023generative}, for example, agents were interviewed about their recent activities, relationships, or future plans, and their answers were cross-checked against internal memory logs. Human evaluators then rated responses for consistency, plausibility, and coherence, providing a qualitative assessment of role fidelity. This procedure can be extended through LLM-based judgments, where a strong LLM serves as the evaluator to assess the quality of agent behaviors, offering scalability and reproducibility. To mitigate bias, a practical solution is to adopt hybrid protocols, where LLM judgments are guided by rubrics defined by human experts and their outputs are validated through human spot-checking.

\section{Conclusion}
\label{sec:conclusion}

This survey provides an up-to-date review of LLMGAs through an analytical framework. At the single agent level, we synthesize prior work across three core components, memory, reasoning, and perception-action interface, that together describe how agents perceive, think, and act through language. Extending this foundation, we introduce a complementary multi-agent framework for analyzing communication protocols and organizational structures that govern coordination, task allocation, and large-scale stability. We further introduce a challenge-centered taxonomy that maps six major game genres to their dominant agent design requirements, from low-latency response in action games to open-ended goal generation in sandbox worlds. Together, these perspectives present a coherent view of how language-enabled agents operate in interactive game environments and outline key challenges that define the next stage of research.

\section*{Acknowledgements}

This research is partially sponsored by the NSF CISE grants 2302720 and 2312758, and a CISCO research grant in Edge AI, and PACE at the Georgia Institute of Technology. The first author and the last author are the primary contacts for this work.

\bibliographystyle{abbrv}
\bibliography{main}

\end{document}